%% file: neural-mte-csl.tex
\documentclass[review]{elsarticle}

\usepackage{hyperref}

\usepackage{url}
\usepackage{latexsym}
\usepackage{graphicx}
\usepackage{color}
\usepackage{booktabs}
\usepackage{subcaption}
\usepackage{multirow}
\usepackage{amsmath}
\usepackage{amssymb}

\input{defs}

\journal{Journal of Computer Speech and Language}

\begin{document}

\begin{frontmatter}

\title{Machine Translation Evaluation with Neural Networks}

\author{Francisco Guzm\'an, Shafiq Joty, Llu\'is M\`arquez 
\hbox{\rm and} Preslav Nakov\\
ALT Research Group\\
Qatar Computing Research Institute --- HBKU, Qatar Foundation\\
{\tt\{fguzman,sjoty,lmarquez,pnakov\}@qf.org.qa}}

\begin{abstract}
We present a framework for machine translation evaluation using neural networks in a pairwise setting, where the goal is to select the better translation from a pair of hypotheses, given the reference translation. In this framework, lexical, syntactic and semantic information from the reference and the two hypotheses is embedded into compact distributed vector representations, and fed into a multi-layer neural network that models nonlinear interactions between each of the hypotheses and the reference, as well as between the two hypotheses. We experiment with the benchmark datasets from the WMT Metrics shared task, on which we obtain the best results published so far, with the basic network configuration. We also perform a series of experiments to analyze and understand the contribution of the different components of the network. We evaluate variants and extensions, including fine-tuning of the semantic embeddings, and sentence-based representations modeled \new{with convolutional and recurrent neural networks}. In summary, the proposed framework is flexible and generalizable, allows for efficient learning and scoring, and provides an MT evaluation metric that correlates with human judgments, and is on par with the state of the art.
\end{abstract}


\end{frontmatter}


\newpage
\section{Introduction}
\label{sec:intro}
\input{introduction}

\section{Related Work}
\label{sec:related}
\input{related}

\section{Pairwise Neural Architecture for MT Evaluation}
\label{sec:architecture}
\input{architecture}

\section{Experiments and Results}
\label{sec:results}
\input{results}

\section{Extensions}
\label{sec:discussion}
\input{discussion-acl2015.1}

\input{discussion-acl2015.2}
\input{discussion-new-paco}

\input{discussion-new-shafiq.2}

\section{An MT Evaluation Metric with Absolute Scores}
\label{sec:absolute-scores}
\input{absolute-scores}
\section{Conclusions and Future Work}
\label{sec:conclusions}
\input{conclusions}


\section*{References}
\bibliographystyle{elsarticle-num}
\bibliography{bibliography/ACL} 

\end{document}

%% file: defs.tex
\DeclareMathOperator \real{\mathbb{R}}

\DeclareMathOperator*{\Ber}{Ber}
\DeclareMathOperator*{\sig}{sig}

\newcommand{\emptytrans}{{t}_\varnothing}

\newcommand{\new}[1]{#1}

\newcommand{\disco}{\nobreak{DiscoTK}}

\newcommand{\spede}{\nobreak{\sc spede07pP}}

\newcommand{\amber}{\nobreak{AMBER}}
\newcommand{\meteor}{\nobreak{\sc Meteor}}

\newcommand{\simpbleu}{\nobreak{SIMPBLEU}}
\newcommand{\ter}{\nobreak{TER}}
\newcommand{\bleu}{\nobreak{BLEU}}

\newcommand{\nist}{\nobreak{NIST}}

\newcommand{\asiya}{\nobreak{\sc Asiya}}

\newcommand{\Ni}{({\em i})~}
\newcommand{\Nii}{({\em ii})~}
\newcommand{\Niii}{({\em iii})~}
\newcommand{\Niv}{({\em iv})~}

\newcommand{\metrics}{\nobreak{\sc 4metrics}}
\newcommand{\syntax}{\nobreak{\sc syntax25}}
\newcommand{\wiki}{\nobreak{\sc Wiki-GW50}}
\newcommand{\wikii}{\nobreak{\sc Wiki-GW300}}
\newcommand{\commoncrawl}{\nobreak{\sc CC-300-42B}}
\newcommand{\commoncrawll}{\nobreak{\sc CC-300-840B}}
\newcommand{\composes}{\nobreak{\sc Composes400}}
\newcommand{\wordvec}{\nobreak{\sc word2vec300}}
\newcommand{\bleucomp}{\nobreak{\sc BLEUcomp}}

%% file: introduction.tex

\noindent Automatic machine translation (MT) evaluation is a necessary step when developing or comparing MT systems. \emph{Reference}-based MT evaluation, i.e., comparing the system output to one or more human reference translations, is the most common approach. Existing MT evaluation measures typically output an absolute quality score by computing the similarity between the machine- and the human-proposed translations.
In the simplest case, the similarity is computed by counting word $n$-gram matches between the translation and the reference. This is the case of \bleu~\cite{Papineni:Roukos:Ward:Zhu:2002}, which has been the standard for MT evaluation for years. Nonetheless, more recent evaluation measures take into account various aspects of linguistic similarity and achieve better correlation with human judgments. 
For instance, synonymy and paraphrasing~\cite{Lavie:2009:MMA}, syntax~\cite{Gimenez2007,Popovic2007,Liu2005}, semantics~\cite{Gimenez2007,Lo2012}, and discourse~\cite{Comelles2010,Wong2012,discoMT:acl2014,discoMT:WMT2014}. The combination of all these aspects led to improved results in metric evaluation campaigns, such as the \emph{WMT Metrics Shared Task}~\cite{WMT-MT14,WMT-MT15}. 

Having quality scores at the sentence level allows ranking alternative translations for a given source sentence. This is useful, for instance, for statistical machine translation (SMT) parameter tuning, for system comparison, and for assessing the progress during MT system development. The quality of automatic MT evaluation metrics is usually determined by computing their correlation with human judgments. To that end, quality rankings of alternative translations have been created by human judges. It is known that assigning an absolute score to a translation is a difficult task for humans. Hence, ranking-based evaluations, where judges are asked to rank the output of \hbox{2 to 5} systems, have been used in recent years, which has yielded much higher inter-annotator agreement \cite{callisonburch-EtAl:2007:WMT}.

These human quality judgments can be used to train automatic metrics. The supervised learning can be oriented to predict absolute scores, e.g., using regression~\cite{albrecht:2008}, or rankings~\cite{Duh:2008,song-cohn:2011:WMT}. A particular case of the latter is used to learn in a pairwise setting, i.e., given a reference and two alternative translations (or hypotheses), the task is to decide which one is better. This setting emulates closely how human judges perform evaluation assessments in reality. From a machine learning perspective, the challenge is to learn, from a pair of hypotheses, which are the features that help to discriminate the better from the worse translation. 

In previous work \cite{guzman-EtAl:2014}, we presented a learning framework for this pairwise setting, based on preference kernels and support vector machines (SVM). We obtained promising results using a combination of syntactic and discourse-based structures. However, using convolution kernels over complex structures comes at a high computational cost both at training and at testing time because the use of kernels requires that the SVM operate in the much slower dual space. Thus, some simplification is needed to make it practical.

While there are some solutions in the kernel-based learning framework to alleviate the computational burden, we took a different direction and presented in~\cite{guzman-EtAl:2015:ACL-IJCNLP} the first neural network (NN) approach for MT evaluation, learning in the pairwise setting. The present article builds on that previous paper and explores some new additions while extending its analysis.

In the core NN model, lexical, syntactic and semantic information from the reference and the two hypotheses is compacted into relatively small distributed vector representations and fed into the input layer, together with a set of individual real-valued features coming from simple pre-existing MT evaluation metrics. 
A hidden layer, motivated by our intuitions on the pairwise ranking problem, is used to capture interactions between the relevant input components. Our evaluation results on the \emph{WMT12 Metrics Shared Task} benchmark datasets~\cite{WMT12} show high correlation with human judgments.
These results clearly surpass~\cite{guzman-EtAl:2014} and are on par with the best results reported for this dataset, achieved by \hbox{DiscoTK~\cite{discoMT:WMT2014}}, which is a much heavier combination metric. 
Interestingly, we empirically show that the syntactic and semantic embeddings produce sizeable and cumulative gains in performance over a strong combination of pre-existing MT evaluation measures (\bleu, \nist, \meteor, and \ter). 

Another advantage of the proposed architecture is efficiency. Due to the vector-based compression of the linguistic structure and the relatively reduced size of the network, testing is fast, which would greatly facilitate the practical use of this approach in real MT evaluation and development. 

In this paper, we broaden the discussion from~\cite{guzman-EtAl:2015:ACL-IJCNLP} by exploring two new model extensions, one oriented to fine-tuning the semantic embeddings on the task data, and the second to produce a sentence-level semantic representation of the input texts based on convolutional and recurrent neural networks. Better results could arguably be obtained by following these approaches, the tradeoff being substantial increase in complexity and reduction in efficiency/speed.

Additionally, we use the pairwise network to produce an absolute quality score when applied to a single input translation, i.e., as a standard MT evaluation metric. 
The pairwise setting is sufficient for most evaluation and MT development scenarios, and we claim that it should be preferred for the cases in which one has to compare a set of hypothesis translations to select the best one (ranking problem). However, one might also need to compare one's 
system to another system on a benchmark dataset, for which one 
knows the evaluation score but not the actual translations. In that case, the comparison requires the use of an evaluation metric that produces an absolute quality score for each system independently. As mentioned before, here we show how the network trained in the pairwise fashion can also be used to produce a high-quality MT evaluation metric over individual translations, which performs comparably to the state of the art both at the sentence and at the system levels.

The rest of the article is organized as follows. Section~\ref{sec:related} overviews the related work.
Section~\ref{sec:architecture} introduces the proposed pairwise NN architecture in its basic form. Section~\ref{sec:results} discusses the experimenal setup, and the results obtained on the benchmark datasets.
Section~\ref{sec:discussion} presents all the variants and extensions of the network mentioned above, together with specific experiments to test their impact. Section~\ref{sec:absolute-scores} discusses the application of the neural network as an evaluation metric for a single translation and compares its results to the state of the art. Finally, Section~\ref{sec:conclusions} concludes and discusses some topics for future research.

%% file: related.tex

\noindent Contemporary MT evaluation measures have evolved beyond simple lexical matching, and now take into account various aspects of linguistic structures, including synonymy and paraphrasing~\cite{Lavie:2009:MMA}, syntax~\cite{Gimenez2007,Popovic2007,Liu2005,stanojevic-simaan:2015:WMT}, semantics~\cite{Gimenez2007,Lo2012}, and even textual entailment~\cite{pado-EtAl:2009:ACLIJCNLP} and discourse relations~\cite{Comelles2010,Wong2012,discoMT:acl2014,discoMT:WMT2014}. The combination of several of these aspects has led to improved results in metric evaluation campaigns, such as the \emph{WMT metrics task} (e.g., \cite{WMT-MT14,WMT-MT15}).

In this paper, we present a general framework for learning from human annotated examples to discriminate better from worse translations. The model uses information from several linguistic representations of the pair of compared translations and the reference. Applying supervised learning to learn or tune MT evaluation metrics is not new. For instance, Kulesza and Shieber~\cite{kulesza2004}, trained an SVM classifier to discriminate good from bad translations, which used lexical and syntactic features, together with other metrics, e.g., BLEU and NIST. Compared to ours, their setting is not a pairwise comparison of two competing translations, but a classification task to distinguish \emph{human}- from \emph{machine-produced} translations. Moreover, in their work, using syntactic features decreased the correlation with human judgments dramatically (although classification accuracy improved), while in our case the effect is positive.

Our learning framework also has connections with the ranking-based approaches for learning to reproduce human judgments of MT quality.
In particular, our setting is similar to that of Duh~\cite{Duh:2008}, but differs from it both in terms of the feature representation and of the learning framework. For instance, we integrate several layers of linguistic information, while Duh \cite{Duh:2008} only used lexical and part-of-speech (PoS) matches as features. Secondly, we use information about both the reference and the two alternative translations simultaneously in a neural-based learning framework capable of modeling complex interactions between the features.

In our previous work \cite{guzman-EtAl:2014}, we introduced a learning framework for the pairwise setting, based on preference kernels and SVMs. We used lexical, PoS, syntactic and discourse-based information in the form of tree-like structures to learn to differentiate better from worse translations. However, in that work we used convolutional kernels, which is computationally expensive and does not scale well to large datasets and complex structures such as graphs and enriched trees. This inefficiency arises both at training and testing time. 
As a main difference, in the present work we use neural embeddings and multi-layer neural networks to train the evaluation metric, which yields an efficient learning framework that works significantly better on the same datasets (although we are not using exactly the same information for learning).

The huge interest in recent years for deep neural nets (NNs) and word embeddings has reached virtually all areas of NLP, in particular, statistical machine translation. For example, in SMT we have observed an increased use of neural nets for language modeling~\cite{Bengio03,Mikolov10} as well as for improving the translation model~\cite{Devlin14,kalchbrenner13,SutskeverVL14}, by creating the so-called \emph{neural machine translation} paradigm. 
However, the application of such models to machine translation evaluation has been much lower. To the best of our knowledge, there are only three independent publications in that direction, which originated in 2015. The first one is our previous paper~\cite{guzman-EtAl:2015:ACL-IJCNLP}, which is the basis for the present article. We adopt the same learning approach and the same core neural network architecture. The novelty in the current article comparing to~\cite{guzman-EtAl:2015:ACL-IJCNLP} is that we explore two significant extensions in the line of improving the semantic representations of the input texts (subsections~\ref{subsec:finetuning} and \ref{subsec:lstms}), and additionally, we show how to use the pairwise architecture to create an MT evaluation metric with absolute scores (Section~\ref{sec:absolute-scores}).

\new{The other two were initially published in WMT in 2015. In~\cite{chen-etal:2015:WMT,chen-guo:2015:ACL-IJCNLP} a metric called {\sc Dreem} is presented, which combined different distributed representations of words and sentences: one-hot, distributed word representations trained with a neural network, and distributed sentence representations learnt with recursive auto-encoders. The vector representations of the translation and the reference were compared using cosine similarity with a length penalty. The results of {\sc Dreem} were moderate at WMT 2015; the metric scored at the middle of the table at the system level (with very good performance on some language pairs), but it scored significantly lower in the segment-level evaluation.}

\new{In the second work from WMT 2015~\cite{gupta-orasan-vangenabith:2015:WMT,gupta-orasan-vangenabith:2015:EMNLP}, authors introduced an MT metric, \textsc{ReVal}, based on dense vector spaces and Tree Long Short Term Memory networks (Tree-LSTM).} The main feature advocated by the authors is its simplicity and resource-lightness, which makes it efficient and appropriate for intensive use, compared to the heavy combination-based state-of-the-art metrics. The metric also got remarkable results at the WMT 2015 Metrics Task~\cite{WMT-MT15}. Compared to our approach, ReVal is trained to reproduce similarity scores between a translation and a reference, while our network is trained by comparing pairs of translation hypotheses. We also explored the use of LSTMs to produce an improved semantic representation of the input sentences, but in our case the LSTM is sequential. Comparatively, we use more information about the input (in the form of syntactic embeddings and some pre-existing MT metrics), but our approach can still be considered efficient compared to the previous state of the art. Finally, regarding the results, while ReVal is good at the system level, it scores below the state of the art at the segment level. According to our evaluation, we almost match ReVal performance at the system level, and we largely outperform ReVal at the segment level. One reason could be the fact that we include more information to learn the metric. Also important is the fact that we learn directly from the pairwise human annotations, while for ReVal, an additional post-processing of the human annotations is required to generate a quality score for each translation to be used as gold-standard annotation. The pairwise learning allows to be closer to the human annotation procedure, and it also permits to integrate into a neural network architecture the interactions between components that reflect our intuitions about MT evaluation.

\new{Overall, using neural networks for MT evaluation remains an under-explored research direction. For example, the 2016 edition of the WMT metrics task \cite{bojar-EtAl:2016:WMT2} did not add much relevant work. The only NN-based metric there was \textsc{Uow.ReVal}, which was the same \textsc{ReVal} that participated in the WMT15 task except for that the LSTM vector dimension in 2016 was 150 instead of 300 in 2015.}


Finally, it is worth noting that the pairwise neural learning approach presented in this paper has been shown to be robust and applicable to other related text--comparison problems. In~\cite{guzman-EtAl:2016:ACL}, a similar network is applied to the problem of ranking answers in community created forums according to their relevance to a given question. In that case, the input consists of the question and two alternative comments, and the network predicts which of the two comments is a more appropriate answer to the given question. The same basic network presented in this paper, with the addition of some lightweight task-specific features, achieved state-of-the-art results in this community question-answering problem.

%% file: architecture.tex

\noindent Our motivation for using neural networks for MT evaluation is twofold. First, to take advantage of their ability to model complex non-linear relationships efficiently. Second, to have a framework that allows for easy incorporation of rich syntactic and semantic representations captured by word embeddings, which are in turn trained using deep learning.
Below, we describe the learning task, and the neural network architecture we propose for it, which was first introduced in~\cite{guzman-EtAl:2015:ACL-IJCNLP}.

\subsection{Learning Task}
\noindent\new{As justified in Section~\ref{sec:intro}, we approach the problem as a pairwise ranking task, to better model the human task when providing the annotations. More precisely, given two translation hypotheses $t_1$ and $t_2$ (and a reference translation $r$), we want to tell which of the two is better.}\footnote{In this work, we do not learn to predict ties, and ties are excluded from our training data.} Thus, we have a binary classification task, which is modeled by the class variable $y$, defined as follows:
\begin{equation}
y = \left\{ 
\begin{array}{c c}
  1 & \mbox{if $t_1$ is better than $t_2$ given $r$}\\
  0 &  \mbox{if $t_1$ is worse than $t_2$ given $r$}\\ \end{array} \right. 
\end{equation}

We model this task using a feed-forward neural network (NN) of the form:
\begin{equation}
p(y|t_1,t_2,r)= \Ber(y|f(t_1,t_2,r))
\end{equation}

\noindent which is a Bernoulli distribution of $y$ with parameter $\sigma = f(t_1,t_2,r)$,
defined as follows:
\begin{equation}
f(t_1,t_2,r)=\sig(\mathbf{w^T_v}\phi(t_1,t_2,r) + b_v) \label{outfunc}
\end{equation}

\noindent where $\sig$ is the sigmoid function, $\phi(x)$ defines the transformations of the input $x$ through the hidden layer, $\mathbf{w_v}$ are the weights from the hidden layer to the output layer, and $b_v$ is a bias term. 

\subsection{Network Architecture}

\begin{figure}[t]
\centering
\includegraphics[width=.78\textwidth]{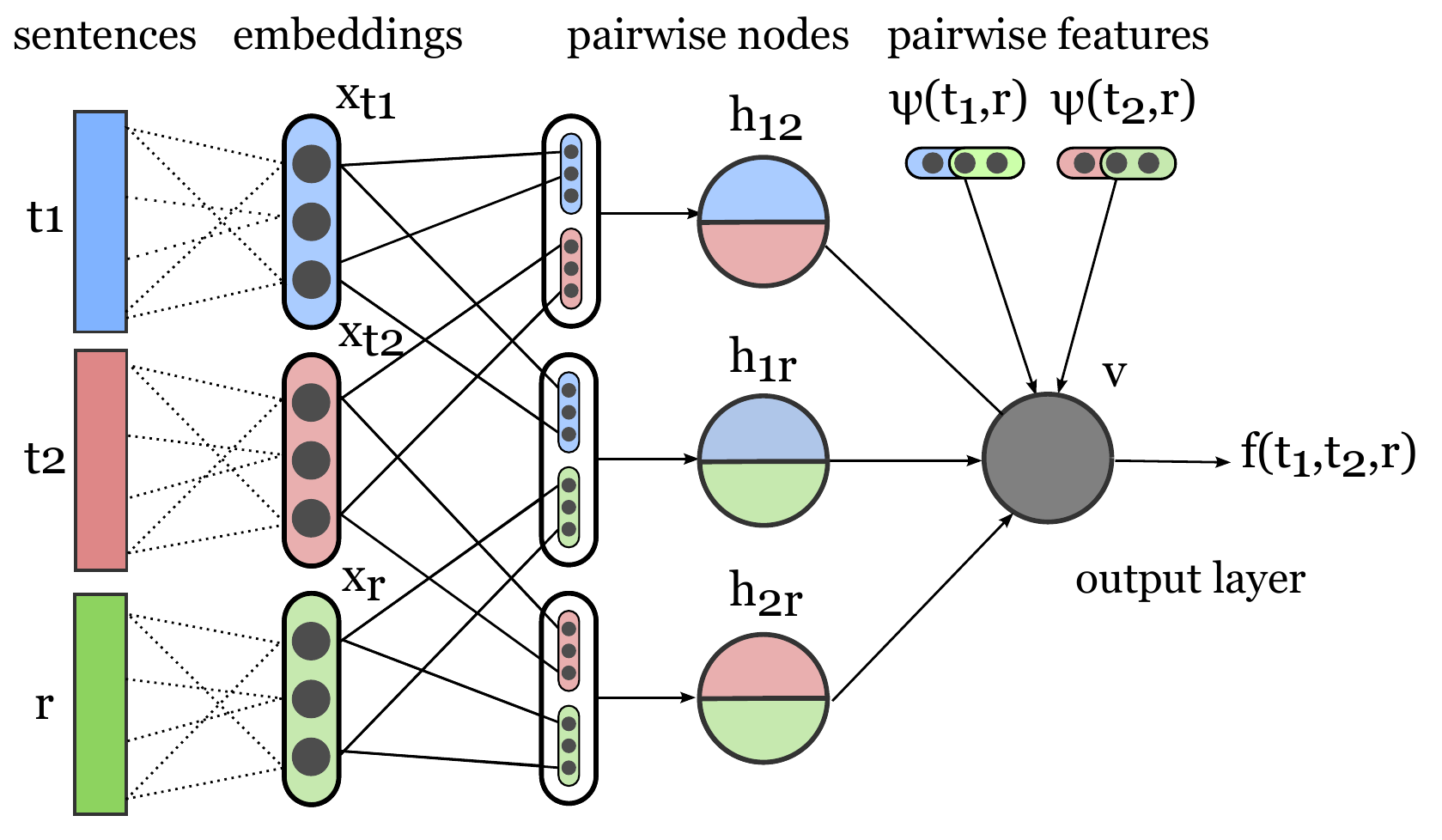}
\caption{\label{fig:architecture} {\small Overall architecture of the neural network.}}
\end{figure}

\noindent In order to decide which hypothesis is \emph{better} given the tuple $(t_1,t_2,r)$ as input, we first map the two hypotheses and the reference to a fixed-length vector $\left[ \mathbf{x}_{t_1},  \mathbf{x}_{t_2}, \mathbf{x}_r \right]$, using syntactic and semantic embeddings. Then, we feed this vector as input to our neural network, whose architecture is shown in Figure~\ref{fig:architecture}. 

In our architecture, we model three types of interactions, using different groups of nodes in the hidden layer. We have two \emph{evaluation} groups $\mathbf{h_{1r}}$ and $\mathbf{h_{2r}}$,
which are inspired by traditional machine translation evaluation metrics
that model how similar each hypothesis $t_i$ is to the reference $r$.

The vector representations of the hypothesis (i.e.,~$\mathbf{x}_{t1}$ or $\mathbf{x}_{t2}$) together with the reference (i.e.,~$\mathbf{x}_{r}$) constitute the input to the hidden nodes in these two groups. The third group of hidden nodes $\mathbf{h_{12}}$, which we call \emph{similarity} group, models how close $t_1$ and $t_2$ are.
This might be useful as highly similar hypotheses are likely to be comparable in quality, irrespective of whether they are good or bad in absolute terms.

The input to each of these groups is represented by concatenating the vector representations of the two components participating in the interaction, i.e., $\mathbf{x_{1r}} = \left[ \mathbf{x}_{t_1}, \mathbf{x}_r \right]$, $\mathbf{x_{2r}} = \left[\mathbf{x}_{t_2}, \mathbf{x}_{r} \right]$, $\mathbf{x_{12}} = \left[ \mathbf{x}_{t_1}, \mathbf{x}_{t_2} \right]$. In summary, the transformation $\phi(t_1,t_2,r) = [\mathbf{h_{12}}, \mathbf{h_{1r}}, \mathbf{h_{2r}}]$ in our NN architecture can be written as follows:   
\begin{eqnarray*}
 \mathbf{h_{1r}} &=& g( \mathbf{W_{1r}} \mathbf{x_{1r}} + \mathbf{b_{1r}})\\
 \mathbf{h_{2r}} &=& g( \mathbf{W_{2r}} \mathbf{x_{2r}} +  \mathbf{b_{2r}})\\
 \mathbf{h_{12}} &=& g( \mathbf{W_{12}} \mathbf{x_{12}} +  \mathbf{b_{12}})
\end{eqnarray*}
\noindent where $g(\cdot)$ is a non-linear activation function (applied component-wise), \hbox{$\mathbf{W}\!\!\in\! \real^{H \times N}$} are the associated weights between the input layer and the hidden layer, and $\mathbf{b}$ are the corresponding bias terms. In our experiments, we used $\tanh$ as an activation function, rather than $\sig$, to be consistent with how parts of our input vectors were generated.\footnote{Many of our input representations consist of word embeddings trained with neural networks that used $\tanh$ as an activation function.} 

In addition, our model allows to incorporate external sources of information by enabling \emph{skip arcs} that go directly from the input to the output, skipping the hidden layer. In our setting, these arcs represent pairwise similarity features between the translation hypotheses and the reference (e.g., the \bleu~ scores of the translations). We denote these pairwise external feature sets as \hbox{$\mathbf{\psi}_{1r}=\psi(t_1,r)$} and  $\mathbf{\psi}_{2r}=\psi(t_2,r)$. When we include the external features in our architecture, the activation at the output, i.e., eq.~(\ref{outfunc}), can be rewritten as follows:  
\begin{equation*}
f(t_1,t_2,r)=\sig(\mathbf{w^T_v} [\phi(t_1,t_2,r), \mathbf{\psi}_{1r}, \mathbf{\psi}_{2r}] + b_v) 
\end{equation*}

\subsection{Network Training}
\label{logcost}

\noindent The negative log likelihood of the training data for the model parameters, \mbox{$\theta = (\mathbf{W_{12}}, \mathbf{W_{1r}}, \mathbf{W_{2r}, w_v}, \mathbf{b_{12}}, \mathbf{b_{1r}}, \mathbf{b_{2r}}, b_v)$,}
can be written as follows:
\begin{equation}
J_{\mathbf{\theta}} = \\
- \sum_{n} y_n \log \hat{y}_{n\theta} + (1-y_n) \log \left(1- \hat{y}_{n\theta} \right) \label{eq:logcost}
\end{equation}

In the above formula, $\hat{y}_{n\theta}=f_n(t_1,t_2,r)$ is the activation at the output layer for the $n$-th data instance. It is also common to use a regularized cost function by adding a weight decay penalty (e.g.,~$L_2$ or $L_1$ regularization) and to perform maximum aposteriori (MAP) estimation of the parameters. We trained our network with stochastic gradient descent (SGD), mini-batches and adagrad updates~\cite{Duchi11}, using Theano~\cite{bergstra+al:2010-scipy}.

%% file: results.tex

\noindent In this section, we first describe the different aspects of our general experimental setup, including the input representations we use to capture the syntactic and semantic features of the two translation hypotheses and the corresponding reference, as well as the datasets used to evaluate the performance of our model. Then we present our first set of results with the basic NN model from Section~\ref{sec:architecture}. In Section~\ref{sec:discussion}, we discuss some variants and extensions of the basic model.

\subsection{Embedding Vectors}

\noindent The embedded representations of the input sentences play a crucial role in our model, since they allow us to model complex relations between the two translations and the reference using syntactic and semantic information.

\paragraph{\bf Syntactic vectors} 
We generate a syntactic vector for each sentence\
using the Stanford neural parser \cite{socher-EtAl:2013:ACL2013},
which generates a 25-dimensional vector as a by-product of syntactic parsing using a recursive NN.
Below we will refer to these vectors as \syntax.

\paragraph{\bf Semantic vectors}
In our basic setting, we compose a semantic vector for a given sentence
using the average of the embedding vectors for the words it contains~\cite{Mitchell:Lapata:2010}.
We use pre-trained, fixed-length word embedding vectors produced by 
\Ni GloVe \cite{pennington-socher-manning:2014:EMNLP2014},
\Nii COMPOSES \cite{P14-1023},
and
\Niii word2vec \cite{mikolov-yih-zweig:2013:NAACL-HLT}.

Our primary representation is based on 50-dimensional GloVe vectors, trained on Wikipedia 2014+Gigaword 5 (6B tokens), to which below we will refer as \wiki. 

In Section~\ref{sec:discussion}, we further experiment with \wikii, the 300-dimensional GloVe vectors trained on the same data, as well as with the \commoncrawl \ and \commoncrawll, 300-dimensional GloVe vectors trained on 42B and on 840B tokens from Common Crawl. We also experiment with the pre-trained, 300-dimensional word2vec embedding vectors, or \wordvec, trained on 100B words from Google News. Finally, we use \composes, the 400-dimensional COMPOSES vectors trained on 2.8 billion tokens from ukWaC, the English Wikipedia, and the British National Corpus.

Finally, also in Section~\ref{sec:discussion} we fine-tune the word embeddings using task supervision, and we also experiment with a recursive representation of the sentences, modeled with LSTMs. 

\subsection{Tuning and Evaluation Datasets}

\noindent We experiment with datasets of segment-level human rankings of system outputs
from the WMT11, WMT12 and WMT13 Metrics shared tasks \cite{WMT11,WMT12,WMT13}.
We focus on translating into English,
for which the WMT11 and the WMT12 datasets can be split by source language:
Czech (cs), German (de), Spanish (es), and French (fr);
WMT13 also has Russian (ru).
There were about 10,000 non-tied human judgments per language pair per dataset.

\subsection{Evaluation Score}

\noindent We evaluate our metrics in terms of correlation with human judgments measured using Kendall's $\tau$. We report $\tau$ for the individual languages as well as macro-averaged across all languages.

Note that there were different versions of $\tau$ at WMT over the years. Prior to 2013, WMT used a strict version, which was later relaxed at WMT13 and further revised at WMT14. See \cite{machacek-bojar:2014:W14-33} for a discussion. Here we use the strict version used at WMT11 and WMT12.

\subsection{Experimental Settings}

\paragraph{Datasets} We train our neural models on WMT11 and we evaluate them on WMT12. 
We further use a random subset of 5,000 examples from WMT13 as a validation set to implement early stopping.

\paragraph{Early stopping} We train on WMT11 for up to 10,000 epochs, and we calculate Kendall's $\tau$ on the development set after each epoch.
We then select the model that achieves the highest $\tau$ on the validation set; in case of ties for the best $\tau$, we select the latest epoch that achieved the highest $\tau$. 

\paragraph{Network parameters} We train our neural network using SGD with adagrad, an initial learning rate of $\eta=0.01$, mini-batches of size $30$, and $L_2$ regularization with a decay parameter $\lambda =1e^{-4}$. We initialize the weights for our matrices by sampling from a uniform distribution following \cite{Xavier10}. We further set the size of each of our pairwise hidden layers $H$ to four nodes, and we normalize the input data using min-max to map the feature values to the range $[-1,1]$.

\subsection{Results}

\begin{table*}[t]
\centering
\vspace{-2mm}
{\footnotesize\begin{tabular}{l@{\hspace*{0.15cm}}l@{\hspace*{0.15cm}}r@{\hspace*{0.15cm}}r@{\hspace*{0.15cm}}r@{\hspace*{0.15cm}}r@{\hspace*{0.1cm}}r}
\toprule
{\bf System} & {\bf Details} & \multicolumn{4}{c}{\bf Kendall's $\tau$} \\\cmidrule(l{2pt}r{2pt}){3-7}

\multicolumn{2}{l}{\bf I. \metrics: commonly-used individual metrics}& \multicolumn{1}{c}{\bf cz} & \multicolumn{1}{c}{\bf de} & \multicolumn{1}{c}{\bf es}  & \multicolumn{1}{c}{\bf fr} & \multicolumn{1}{c}{\bf AVG}\\
\midrule

\bleu & no learning & 15.88 & 18.56 & 18.57 & 20.83 & 18.46\\
\nist & no learning & 19.66 & 23.09 & 20.41 & 22.21 & 21.34\\
\ter & no learning & 17.80 & 25.31& 22.86 & 21.05 & 21.75\\
\meteor & no learning & 20.82 &26.79 & 23.81 & 22.93 & 23.59\\\vspace{-2mm}
\\

\multicolumn{2}{l}{\bf II. NN using syn. and sem. embedding vectors}\\
\midrule
\syntax & multi-layer NN & 8.00 & 13.03 & 12.11 & 7.42 & 10.14 \\
\wiki & multi-layer NN & 14.31 & 11.49 & 9.24 & 4.99 & 10.01 \\\vspace{-2mm}
\\
\multicolumn{2}{l}{\bf III. NN using \metrics\ and embedding vectors}\\
\midrule
\metrics &  logistic regression & 23.46 & 29.95 & 27.49 & 27.36 & 27.06\\
\metrics+\syntax & multi-layer NN & 26.09 & 30.58 & 29.30 & 28.07 & 28.51 \\
\metrics+\wiki & multi-layer NN & 25.67 & 32.50 & 29.21 & {\bf 28.92} & 29.07 \\
\metrics+\syntax+\wiki & multi-layer NN & {\bf 26.30} & {\bf 33.19} & {\bf 30.38} & {\bf 28.92} & {\bf 29.70} \\
\bottomrule 
\end{tabular}}
\caption{\label{t:results}{\small Kendall's tau ($\tau$) on the WMT12 dataset for various metrics. `AVG' is the average $\tau$ for the four language pairs. The best results are marked in boldface.}}
\end{table*}

\noindent The main findings of our experiments are shown in Table~\ref{t:results}.
Section I of Table~\ref{t:results} shows the results for four commonly-used metrics for MT evaluation that compare a translation hypothesis to the reference(s) using primarily lexical information like word and $n$-gram overlap (even though some allow paraphrases): \bleu, \nist, \ter, and \meteor~\cite{Papineni:Roukos:Ward:Zhu:2002,Doddington:2002:AEM,Snover06astudy,Denkowski2011}.
We will refer to the set of these four metrics as \metrics.
These metrics are not tuned and achieve Kendall's $\tau$ between 18.5 and 23.5. These are the metrics that are added as pairwise similarity features in our neural network approach (skip arcs).

Section II of Table~\ref{t:results} shows the results of the multi-layer neural network trained on vectors from word embeddings only: \syntax \ and \wiki.
These networks achieve modest $\tau$ values around 10, which should not be surprising: they use very general vector representations and have no access to word or $n$-gram overlap or to length information, which are very important features to compute similarity against the reference. However, as will be discussed below, their contribution is complementary to the four previous evaluation metrics and will lead to significant improvements in combination with them.

Section III of Table~\ref{t:results} shows the results for neural network setups that combine the four metrics from \metrics \ with \syntax \ and \wiki.
We can see that just combining the four metrics in a flat neural net (i.e., no hidden layer), which is equivalent to logistic regression,
yields a $\tau$ of 27.06, which is better than the best of the four metrics by 3.5 points absolute,
and also better by over 1.5 points absolute than the best metric that participated at the WMT12 metrics task competition (\spede\ with $\tau=25.4$)~\cite{WMT-MT14}.
Indeed, \metrics \ is a strong mix that involves not only simple lexical overlap but also approximate matching, paraphrases, edit distance, lengths, etc.
Yet, adding to \metrics \ the embedding vectors yields sizeable further improvements: +1.5 and +2.0 points absolute when adding \syntax \ and \wiki, respectively.
Finally, adding both yields even further improvements close to $\tau$ of 30 (+2.64 $\tau$ points), showing that lexical semantics and syntactic representations are complementary.

In Section~\ref{sec:absolute-scores} we include a comparison of our results to the state of the art on the same dataset. To provide now some context to our scores from Table~\ref{t:results}, the official evaluation for the top three systems that participated at WMT12, showed values of $\tau$ between 22.9 and 25.4, and the best published result on this dataset is $\tau=30.5$.

%% file: discussion-acl2015.1.tex

\noindent In this section, we explore how different parts of our framework can be modified to improve its performance, or how it can be extended for further generalization. First, we explore variations of the feature sets from the perspective of both the pairwise features and the embeddings (Subsections~\ref{subsec:bleucomp} and \ref{subsec:embeddings}). Then, we analyze the role of the network architecture and of the cost function used for learning (Subsections~\ref{subsec:hidden} and \ref{subsec:costfunction}). Finally, we explore a task-specific fine tuning of the semantic embeddings, and a sentence-based representation of the semantic embeddings based on LSTMs (Subsections~\ref{subsec:finetuning} and \ref{subsec:lstms}).

\subsection{Fine-Grained Pairwise Features}
\label{subsec:bleucomp}

\begin{table*}[t]
\centering
\vspace{-2mm}
\hspace*{-2mm}
{\footnotesize\begin{tabular}{l@{\hspace*{2mm}}l@{\hspace*{2mm}}c@{\hspace*{2mm}}c@{\hspace*{2mm}}c@{\hspace*{2mm}}c@{\hspace*{2.5mm}}c}
\toprule
 &  & \multicolumn{5}{c}{\bf Kendall's $\tau$} \\\cmidrule(l{2pt}r{2pt}){3-7}
\multicolumn{1}{c}{\bf System} & \multicolumn{1}{c}{\bf Details} & { \bf cz }&{\bf de} & {\bf es}  & {\bf fr} &{\bf AVG}\\
\midrule
\bleu & no learning & 15.88 & 18.56 & 18.57 & 20.83 & 18.46\\
\midrule
\bleucomp & logistic regression & 18.18 & 21.13 & 19.79 & 19.91 & 19.75\\
\bleucomp+\syntax & multi-layer NN & 20.75 & 25.32 & 24.85 & 23.88 & 23.70\\
\bleucomp+\wiki & multi-layer NN & 22.96 & 26.63 & 25.99 & 24.10 & 24.92\\
\bleucomp+\syntax+\wiki & multi-layer NN & 22.84 & 28.92 & 27.95 & 24.90 & \bf 26.15\\
\midrule 
{\it \bleu+\syntax+\wiki} & \it multi-layer NN & \it 20.03 & \it 25.95 & \it 27.07 & \it 23.16 & \it 24.05\\
\bottomrule
\end{tabular}}
\caption{\label{t:results:BLEUcomp}{\small Kendall's $\tau$ on WMT12 for neural networks using \bleucomp, a decomposed version of \bleu. For comparison, the last line shows a combination using \bleu \ instead of \bleucomp.} }
\end{table*}

\noindent We have shown that our NN can integrate syntactic and semantic vectors with scores from other metrics. In fact, ours is a more general framework, where one can integrate the \emph{components of a metric} instead of its score, which could yield better learning. Below, we demonstrate this for \bleu.

\bleu \
has different components: 
the $n$-gram precisions,
the $n$-gram matches,
the total number of $n$-grams ($n$=1,2,3,4),
the lengths of the hypotheses and of the reference,
the length ratio between them,
and \bleu's brevity penalty.
We will refer to this decomposed \bleu \ as \bleucomp.
Some of these features were previously used in \simpbleu~\cite{song-cohn:2011:WMT}.

The results of using the components of \bleucomp \ as features are shown in Table~\ref{t:results:BLEUcomp}.
We see that using a single-layer neural network, which is equivalent to logistic regression, outperforms \bleu\ by more than +1.3 $\tau$ points absolute.

As before, adding \syntax\ and \wiki\ improves the results, but now by a more sizable margin: +4 for the former and +5 for the latter. Adding both yields +6.5 improvement over \bleucomp, and almost 8 points over \bleu.

We see once again that the syntactic and semantic word embeddings are complementary to the information sources used by metrics such as \bleu, and that our framework can learn from richer pairwise feature sets such as \bleucomp. 
Moreover, the last line of the table shows that using the fine-grained components of \bleu\ has additive improvements to the combination (+2.1 $\tau$ points over the \bleu-based combination), which suggests that it is better to use as input the components of a metric rather than the metric score.


\subsection{Larger Semantic Vectors}
\label{subsec:embeddings}

\noindent One interesting aspect to explore is the effect of the dimensionality of the input embeddings. Here, we studied the impact of using semantic vectors of bigger sizes, trained on different and larger text collections. The results are shown in Table~\ref{t:results:larger}.
We can see that, compared to the 50-dimensional \wiki,
300-400 dimensional vectors are generally better by 1-2 $\tau$ points absolute when used in isolation;
however, when used in combination with \metrics+\syntax, they do not offer much gain (up to +0.2),
and  in some cases, we observe a slight drop in performance. We suspect that the variability across the different collections is due to a domain mismatch. Yet, we defer this question for future work.

\begin{table}[t]
\centering
{\footnotesize\begin{tabular}{lcc}
\toprule
{ \bf Source }& { \bf Alone}  & { \bf Comb.}\\
\midrule
\wiki         & \it 10.01 & \it 29.70\\
\wikii        &  9.66 & \bf 29.90\\
\commoncrawl  & \bf 12.16 & 29.68\\
\commoncrawll & \bf 11.41 & \bf 29.88\\
\midrule
\wordvec      &  7.72 & 29.13\\
\midrule
\composes     & \bf 12.35 & 28.54\\
\bottomrule
\end{tabular}}
\caption{\label{t:results:larger}{\small Average Kendall's $\tau$ on WMT12 for semantic vectors
trained on different text collections.
Shown are results \Ni when using the semantic vectors alone,
and \Nii when combining them with \metrics \ and \syntax.
The improvements over \wiki \ are marked in bold.}}
\end{table}

%% file: discussion-acl2015.2.tex

\subsection{Deep vs. Flat Neural Network}
\label{subsec:hidden}

\noindent One interesting question is how much of the learning is due to the rich input representations,
and how much happens because of the architecture of the neural network.
To answer this, we experimented with two settings: a single-layer neural network, where all input features are fed directly to the output layer (which is logistic regression), and our proposed multi-layer neural network.

The results are shown in Table~\ref{t:architectures}.
We can see that switching from our multi-layer architecture to a single-layer one yields an absolute drop of 0.6 $\tau$. This suggests that there is value in using the deeper, pairwise layer architecture.

\new{\subsection{Pairwise vs. Fully-connected Neural Network}}
\label{subsec:pairwise:fully}

\noindent \new{Another interesting aspect is how our pairwise neural network compares to a fully connected architecture, where there are connections between each node in the input layer to each node in the hidden layer. A fully connected architecture has a higher number of parameters and is more expressive. However, as the results in Table~\ref{t:architectures} show (compare the last two rows), it does not really yield improvements over our pairwise model. This suggests that our model is expressive enough and captures the interactions that are worth modeling, while leaving out those that are not really needed.}

\begin{table}[tb]
\centering
\small\begin{tabular}{lccccc}
\toprule
 & \multicolumn{5}{c}{\bf Kendall's $\tau$} \\\cmidrule(l{2pt}r{2pt}){2-6}
\multicolumn{1}{c}{\bf Details} & { \bf cz }&{\bf de} & {\bf es}  & {\bf fr} &{\bf AVG}\\
\midrule
 single-layer & 25.86 & 32.06 & 30.03& 28.45 &  29.10 \\
 multi-layer, pairwise & {\bf 26.30} & {\bf 33.19} & {\bf 30.38} & {\bf 28.92} & {\bf 29.70} \\
 \new{multi-layer, fully-connected} & \new{\bf 26.30} & \new{\bf 33.31} & \new{\bf 30.40} & \new{\bf 28.82} & \new{\bf 29.73} \\
\bottomrule
\end{tabular}
\caption{{\small\label{t:architectures} Kendall's tau ($\tau$) on the WMT12 dataset for alternative architectures using \metrics+\syntax+\wiki ~as input.} }
\vspace{-0.25cm}
\end{table}


\subsection{Task-Specific Cost Function}
\label{subsec:costfunction}

\noindent Another question is whether the log-likelihood cost function $J(\theta)$ (see Section~\ref{logcost}) is the most appropriate for our ranking task, provided that it is evaluated using Kendall's $\tau$ as defined below:
\begin{equation}
\tau = \frac{concord. - disc. - ties}{concord+disc. + ties}
\end{equation}

\noindent where \emph{concord.}, \emph{disc.} and \emph{ties} are the number of concordant, disconcordant and tied pairs. 

Given an input tuple $(t_1,t_2,r)$, the logistic cost function yields larger values of $\sigma = f(t_1,t_2,r)$ if $y=1$, and smaller if $y=0$,
where $0 \le \sigma \le 1$ is the parameter of the Bernoulli distribution. However, it does not model \emph{directly} the probability when the order of the hypotheses in the tuple is reversed, i.e., $\sigma'=f(t_2,t_1,r)$. 

For our specific task, given an input tuple $(t_1,t_2,r)$, we want to make sure that the difference between the two output activations $\Delta = \sigma - \sigma'$ is positive when $y=1$, and negative when $y=0$. 
Ensuring this would take us closer to the actual objective, which is Kendall's $\tau$. One possible way to do this is to introduce a task-specific cost function that penalizes the disagreements
similarly to the way Kendall's $\tau$ does.\footnote{Other variations for ranking tasks are possible, e.g.,~\cite{yih-EtAl:2011:CoNLL}.} 
 In particular, we define a new \emph{Kendall cost} as follows:
\begin{equation}
J_{\mathbf{\theta}} = - \sum_{n} y_n\sig(-\gamma \Delta_n) + (1-y_n)\sig (\gamma \Delta_n)\,
\end{equation}

\noindent where we use the sigmoid function $\sig$ as a differentiable approximation to the step function.

The above cost function penalizes disconcordances, i.e., cases where \Ni $y=1$ but $\Delta <0$, or \Nii when $y=0$ but $\Delta>0$.
However, we also need to make sure that we discourage \emph{ties}.
We do so by adding a zero-mean Gaussian regularization term $\exp(-\beta\Delta^2/2)$ that penalizes the value of $\Delta$ getting close to zero.
Note that the specific values for $\gamma$ and $\beta$ are not really important, as long as they are large.
In particular, in our experiments, we used $\gamma = \beta = 100$. 

\begin{table}[t]
\centering
\vspace{-4mm}
\small\begin{tabular}{lccccc}
\toprule
& \multicolumn{5}{c}{\bf Kendall's $\tau$} \\\cmidrule(l{2pt}r{2pt}){2-6}
 \multicolumn{1}{c}{\bf Details} & { \bf cz }&{\bf de} & {\bf es}  & {\bf fr} &{\bf AVG}\\
\midrule
 Logistic & 26.30 & 33.19 & 30.38& 28.92 & 29.70 \\
 Kendall  & 27.04 & 33.60 & 29.48& 28.54 & 29.53 \\
 Log.+Ken.  & 26.90 & 33.17& 30.40& 29.21 & {\bf 29.92}\\
\bottomrule
\end{tabular}
\caption{{\small\label{t:costfun} Kendall's tau ($\tau$) on  WMT12 for alternative cost functions using \metrics+\syntax+\wiki.} }
\vspace{-0.35cm}
\end{table}


Table~\ref{t:costfun} shows a comparison of the two cost functions: \Ni the standard logistic cost, and \Nii our Kendall cost.
We can see that using the Kendall cost enables effective learning, although it is eventually outperformed by the logistic cost.
Our investigation revealed that this was due to a combination of slower convergence and poor initialization.
Therefore, we further experimented with a setup where we first used the logistic cost to pre-train the neural network,
and then we switched to the Kendall cost in order to perform some finer tuning.
As we can see in Table~\ref{t:costfun} (last row), doing so yielded a sizable improvement over using the Kendall cost only;
it also improved over using the logistic cost only.
 
%


%% file: discussion-new-paco.tex

\subsection{Fine-tuning of the embedded representations}
\label{subsec:finetuning}

\noindent In our experiments so far, we have used \emph{fixed} semantic word-embedding representations. These were pre-computed and used as features in our network. In this section, we fine-tune the word embedding matrix to produce task-specific sentence-level representations using the feedback from our task. 

We represent each word in the vocabulary $V$ by a $D$ dimensional vector in a shared embedding matrix $E$ $\in$ $\real^{|V| \times D}$; $E$ is considered a model parameter to learn. We can initialize $E$ randomly or with pretrained word embedding vectors like \new{word2vec \cite{mikolov2013efficient} or Glove \cite{pennington-socher-manning:2014:EMNLP2014}}.

Given an input sentence $\mathbf{s} = (w_1, \cdots, w_T)$, we first transform it into a feature sequence by mapping each token $w_t \in \mathbf{s}$ to a one-hot vector $\mathbf{f}_t$, and generate an input vector $\mathbf{x}_t:E^T\mathbf{f}_t \in\real^{D}$ for each token $w_t$. Then, we produce a semantic representation for the sentence sentence by \emph{averaging} the embeddings. This is equivalent to computing the dot product between the embedding matrix $E$ and the one-hot vector $\mathbf{f}$ for the whole sentence and divide it by the number of words in the sentence: $\mathbf{x}=\frac{1}{N}E^T\mathbf{f}$.

\paragraph{Normalization issues} A first complication that arises from using word embeddings to compose sentence-level representations \emph{on the fly}, as opposed to pre-computed representations, stems from normalization. When using pre-computed sentence-level embedding features, we can enforce sentence-level embedding coefficients in each dimension to be restricted to the range $[-1,1]$  by using min-max normalization. To do so, we determine min-max parameters for each of the dimensions of the translations and the references independently, using the training data. However, when using sentence-level embeddings composed \emph{on the fly}, normalizing the data is not trivial because the parameters of the embedding matrix $E$ are now shared between translations and references, thus making it infeasible to reproduce the same normalized sentence-level representations as before. 

Therefore, below we first study the drop in performance due to the lack of normalization by 
comparing the results of our full system from Section~\ref{sec:results} 
to the same system when using sentence embeddings computed on the fly, with no fine-tuning. Then, we calculate the improvements obtained by fine-tuning the word embeddings with the task-specific feedback.

\paragraph{Fine-tuning} Learning high-quality word embeddings requires a lot of mono-lingual data to make correct estimations. Therefore, here we use pre-trained  \wiki\ word embeddings to initialize our word-embedding matrices. Our learning task is thus limited to fine-tuning the embedding matrix to produce task-specific sentence representations. To measure the effect of learning these task-specific representations, we experiment with two different settings: First, we use a \emph{moderate} fine tuning, in which 
we introduce a regularization term that penalizes large deviations from the initialization matrix, $E^0$. In other words, the regularization term is proportional to $\sum (E_{ij} - E^0_{ij})^2$.
Additionally, we use the \emph{full} version of fine-tuning. In this second case, $E^0$ is only used as an initialization for the learning process, and the matrix $E$ is allowed to update freely, only constrained by the $L_2$ regularization, i.e., $\sum E_{ij}^2$.

\paragraph{Results} In Table~\ref{tab:discussion}, we observe the results on the WMT12 dataset measured by Kendall's $\tau$ in the different settings described above. First, note that the effect of normalization is noticeable. Just by switching to a dynamic composition of sentence-level embeddings, we lose $0.16$ points absolute. Allowing the moderate fine-tuning of the embedding matrix only improves performance slightly by $0.07$. However, allowing the full fine-tuning of the embedding matrix, yields an improvement of $0.25$ over the un-tuned setting, and even slightly over the fully normalized baseline system (+$0.09$).

These results suggest that using task-specific embedding representations is useful and can lead to sizeable gains in performance. In our experiments, these improvements in performance come from better word embeddings that depart substantially from the original pre-computed embeddings. This is encouraging, as it confirms that task-specific representations perform better than generic ones. 

However, there is a tradeoff: by computing sentence-level representations on the fly, we lose the benefits of feature normalization, which in our setting leads to a substantial drop in performance. Furthermore, the increase in computational complexity that happens by computing sentence-level representations on the fly makes learning around 30 times slower,\footnote{Note also that there is a tradeoff between space and time complexity between these two approaches. Pre-computing sentence-level vectors is less efficient in terms of disk space.} which makes it hard to justify by the slight increase of performance with respect to the baseline system with pre-computed embeddings.

\begin{table*}[t]
\centering
\vspace{-2mm}
{\footnotesize\begin{tabular}{l@{\hspace*{0.2cm}}c@{\hspace*{0.15cm}}c@{\hspace*{0.15cm}}c@{\hspace*{0.15cm}}c@{\hspace*{0.2cm}}c}
\toprule
   & \multicolumn{5}{c}{\bf Kendall's $\tau$} \\\cmidrule(l{2pt}r{2pt}){2-6}
\multicolumn{1}{c}{\bf Details} & {\bf cz }&{\bf de} & {\bf es}  & {\bf fr} & {\bf AVG}\\
\midrule
Pre-computed sentence embeddings 
& 26.25  & 33.58&  30.67&  28.40&  29.72\\
\midrule
On-the-fly sentence embeddings, no fine-tuning &26.25&  33.78&  30.32&  27.89& 29.56 \\
On-the-fly sentence embeddings + moderate fine-tuning & 26.50&  33.64&  30.01&  28.35&  29.63\\
On-the-fly sentence embeddings + full fine-tuning & 26.92&  33.69&  30.11&  28.51&29.81 \\
\bottomrule
\end{tabular}}
\caption{\label{tab:discussion}{\small Kendall's $\tau$ on WMT12 for neural networks using different variants of word-embedding fine tuning. All variants are implemented on top of our full system from Section~\ref{sec:results}, referred to as \emph{``Pre-computed sentence embeddings''} in this table.}}
\end{table*}

%% file: discussion-new-shafiq.2.tex

\subsection{Sentence-based representation of input texts}
\label{subsec:lstms}

\noindent One aspect in which our proposed model is extremely simple is that it computes the semantic representation of a sentence by just \emph{averaging} the embedding vectors of its words. \new{In this continuous bag-of-words (BOW) approach, we do not model any local or global structure of the sentence. However, capturing phrasal structures and their compositionality could be important for distinguishing a better translation from a worse one. Thus, below we explore Convolutional Neural Networks (CNN) \cite{kim:2014:EMNLP2014} to encode local phrasal structures and Recurrent Neural Networks (RNN) with a Long Short Term Memory (LSTM) hidden layer~\cite{hochreiter1997long} to encode the global structure of a sentence, and to fine-tune the word vectors simultaneously.}

\begin{figure*}[t!]
\centering
\scalebox{0.60}{
  \centering
  \includegraphics[width=1\linewidth]{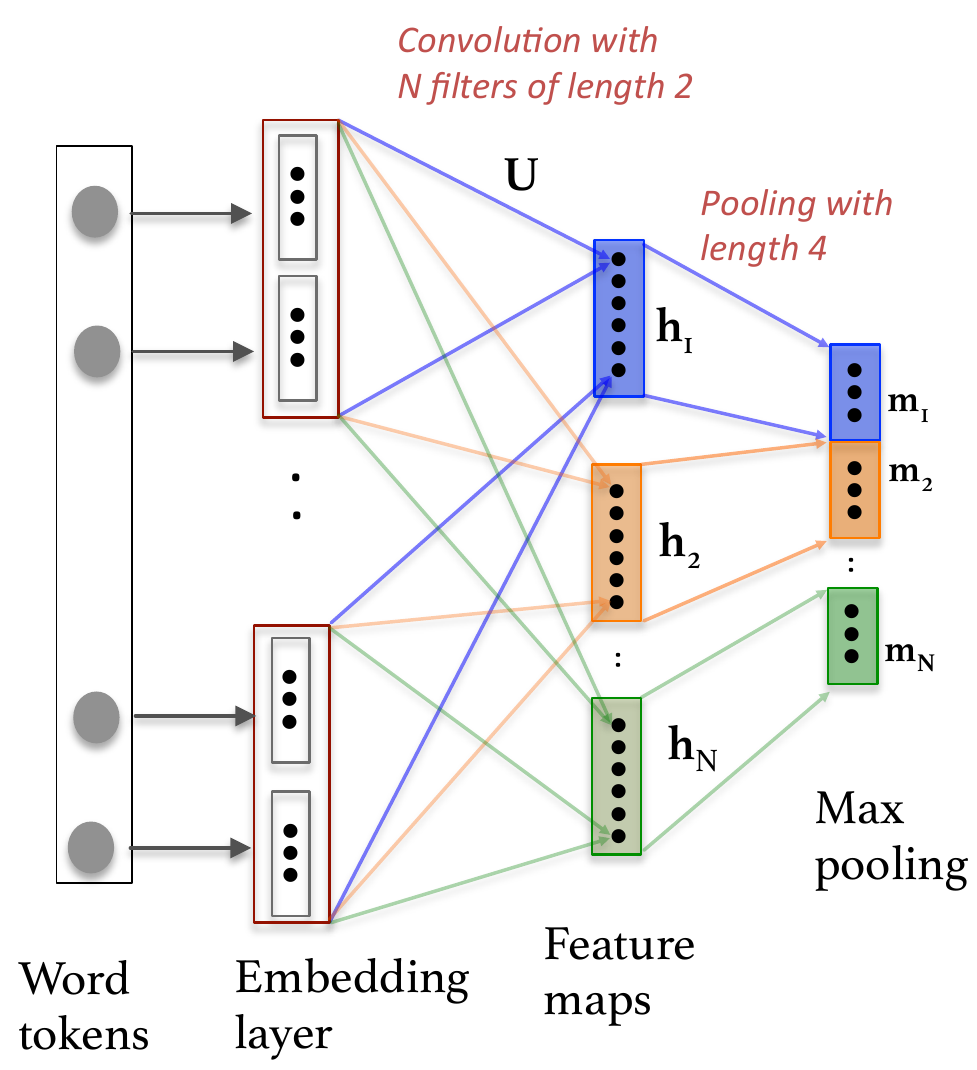}
  \label{fig:sfig1}}
\caption{Convolutional neural network for sentence representation.}
\label{fig:cnn}
\end{figure*}

\new{\subsubsection{Convolutional Neural Network} \label{sec:cnn}
Figure \ref{fig:cnn} demonstrates how our CNN encodes a sentence into a fixed-length vector by means of \emph{convolution} and \emph{pooling} operations. Similar to the fine-tuning setting discussed above, each word token $w_t \in \mathbf{s}$ is first mapped into a vector $\mathbf{x_t}\in\real^{D}$ by looking up the  embedding matrix $E$. The vectors are then passed through a sequence of convolution and pooling operations, which yields a high-level abstract representation of the sentence.}

\new{A convolution operation involves applying a \emph{filter} $\mathbf{u} \in \real^{L.D}$ to a window of $L$ words to produce a new feature}  

\begin{equation}
\new{h_t = g(\mathbf{u} . \mathbf{x}_{t:t+L-1} + b_t)}
\end{equation}

\new{\noindent where $\mathbf{x}_{t:t+L-1}$ denotes the concatenation of $L$ input vectors, $b_t$ is a bias term, and $g$ is a nonlinear activation function. We apply this filter to each possible $L$-word window in the sentence to generate a \emph{feature map} $\mathbf{h}_i = [h_1, \cdots, h_{T+L-1}]$. We repeat this process $N$ times with $N$ different filters to get $N$ different feature maps. We use a \emph{wide} convolution \cite{Kalchbrenner14} (as opposed to \emph{narrow}), which ensures that the filters reach the entire sentence, including the boundary words. This is done by performing \emph{zero-padding}, where out-of-range (i.e., $t$$<$$1$ or $t$$>$$T$) vectors are assumed to be zero.}


\new{After convolution, we apply a max-pooling operation to each feature map:}

\begin{equation}
\new{\mathbf{m} = [\mu_p(\mathbf{h}_1), \cdots, \mu_p(\mathbf{h}_N)]} \label{max_pool}
\end{equation}
  
\new{\noindent where $\mu_p(\mathbf{h}_i)$ refers to the $\max$ operation applied to each window of $p$ features in the feature map $\mathbf{h}_i$. For instance, with $p=2$, this pooling gives the same number of features as in the feature map (because of the zero-padding).}

\new{Intuitively, the filters compose local $n$-grams into higher-level representations in the feature maps, and max-pooling reduces the output's dimensionality while keeping the most important aspects from each feature map. This design of CNNs yields fewer parameters than its fully-connected counterpart, and thus generalizes well for target prediction tasks. Since each convolution-pooling operation is performed independently, the features extracted become invariant of location (i.e., where they occur in the sentence), and act like bag-of-$n$-grams. 
However, capturing long-range structural information could be important for modeling sentences. Thus, below we further describe an LSTM-RNN architecture that capture long-range structural information.}

\subsubsection{Long Short Term Memory Recurrent Neural Network}

\begin{figure*}[tb!]
\scalebox{0.98}{
\begin{subfigure}{.60\textwidth}
  \centering
  \includegraphics[width=1\linewidth]{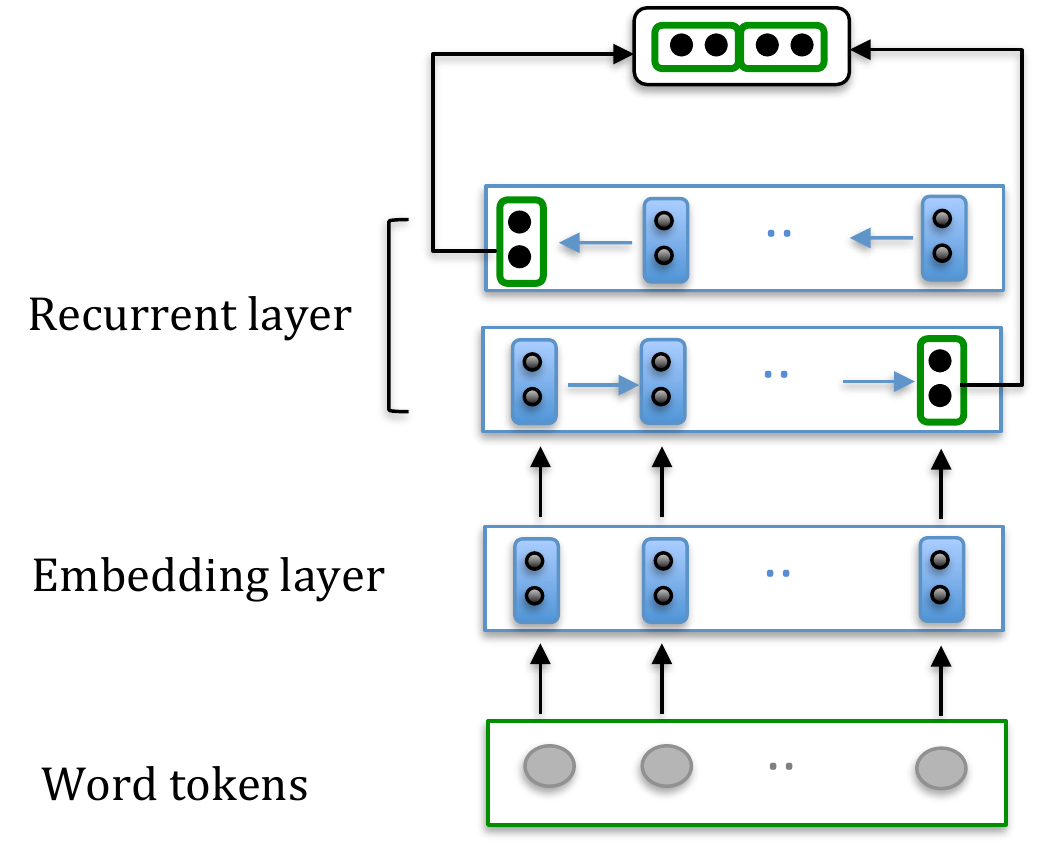}
  \caption{Bidirectional LSTM for sentence representation}
  \label{fig:sfig1}
\end{subfigure}
\begin{subfigure}{.40\textwidth}
  \centering
  \includegraphics[width=1\linewidth]{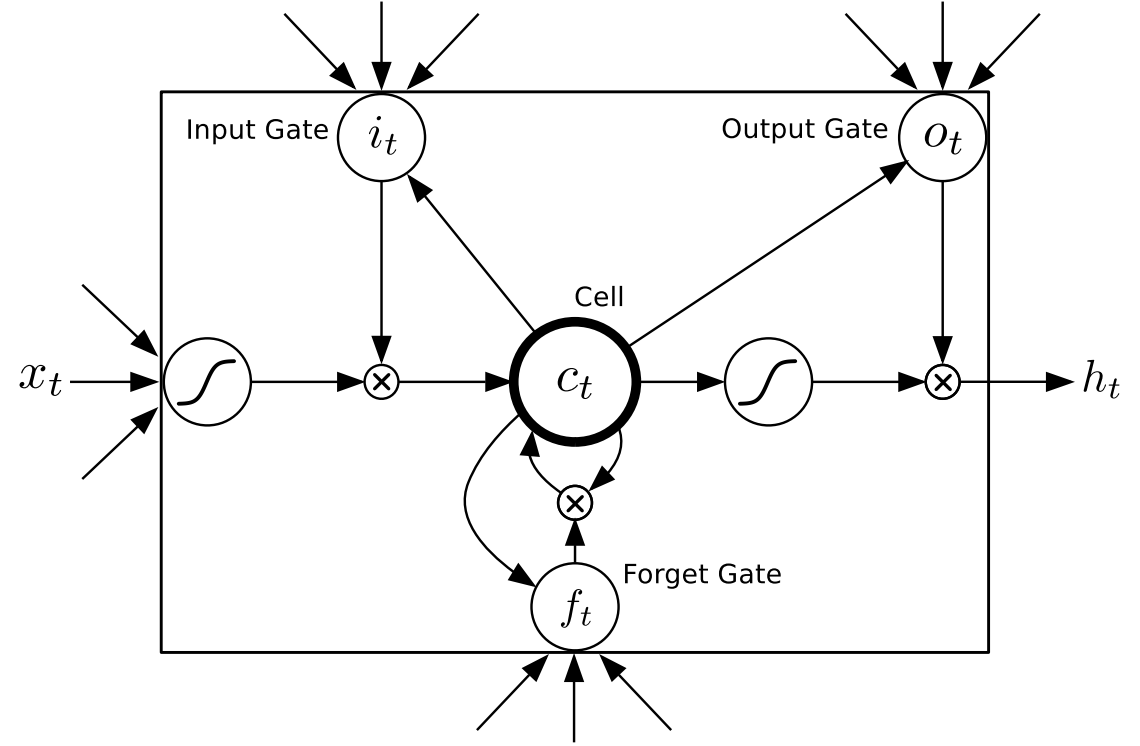}
  \caption{An LSTM cell}
  \label{fig:sfig2}
\end{subfigure}}
\caption{LSTM-based recurrent neural network for sentence representation}
\label{fig:lstm-framework}
\end{figure*}

\new{RNNs encode a sentence into a vector by processing its words sequentially, at each time step combining the current input with the previous hidden state (Figure~\ref{fig:lstm-framework}a).  We experiment with both unidirectional and bidirectional RNNs. }

\new{In this setting, after mapping each word token to its embedding vector in $E$, the vector is passed} to the LSTM recurrent layer, which computes a compositional representation $\overrightarrow{\mathbf{h}}_t$ at every time step $t$ by performing nonlinear transformations of the current input  $\mathbf{x}_t$ and the output of the previous time step $\overrightarrow{\mathbf{h}}_{t-1}$. Specifically, the recurrent layer in an LSTM-RNN is formed by hidden units called \emph{memory blocks}. A memory block is composed of four elements: \Ni a memory cell $c$ (a neuron) with a self-connection, \Nii an input gate $i$ to control the flow of input signal into the neuron, \Niii an output gate $o$ to control the effect of the neuron activation on other neurons, and \Niv a forget gate $f$ to allow the neuron to adaptively reset its current state through the self-connection. The following sequence of equations describe how the memory blocks are updated at every time step $t$:        


\begin{eqnarray}
	\mathbf{i}_t &=& \sig(U_i\mathbf{h}_{t-1} + V_i\mathbf{x}_t + \mathbf{b}_i)  \label {lstm_first}\\
    \mathbf{f}_t &=& \sig(U_f\mathbf{h}_{t-1} + V_f\mathbf{x}_t + \mathbf{b}_f) \\
    \mathbf{c}_t &=& \hspace{-0.1cm}  \mathbf{i}_t\odot \tanh(U_c\mathbf{h}_{t-1} + V_c\mathbf{x}_t) + \mathbf{f}_t\odot\mathbf{c}_{t-1} \\
    \mathbf{o}_t &=& \sig(U_o\mathbf{h}_{t-1} + V_o\mathbf{x}_t + \mathbf{b}_o) \\
    \mathbf{h}_t &=& \mathbf{o}_t \odot \tanh(\mathbf{c}_t) \label {lstm_last}
\end{eqnarray}

\noindent where $U_k$ and $V_k$ are the weight matrices between two consecutive hidden layers, and between the input and the hidden layers, respectively, which are associated with gate $k\in$ \{input, output, forget, cell\}; and $\mathbf{b}_k$ is the corresponding bias vector. The symbols $\sig$ and $\tanh$ denote hard sigmoid and hyperbolic tangent, respectively, and the symbol $\odot$ denotes an element-wise product of two vectors.
 
LSTM, by means of its specifically designed gates (as opposed to simple RNNs), is capable of capturing long-distance dependencies. We can interpret $\mathbf{h}_t$ as an intermediate representation summarizing the past. The output of the last time step $\overrightarrow{\mathbf{h}_T}$ thus represents the whole sentence, which can be fed to the subsequent layers of the neural network architecture. 

\paragraph{Bidirectionality} 
The RNN described above encodes information from the past only. However, information from the future could also be crucial, especially for longer sentences, where a unidirectional RNN can be limited in encoding the necessary information into a single vector. Bidirectional RNNs \cite{Schuster:1997} capture dependencies from both directions, thus providing two different views of the same sentence. This amounts to having a backward counterpart for each of the equations from \ref{lstm_first} to \ref{lstm_last}. Each sentence in a bidirectional LSTM-RNN is thus represented by the concatenated vector $[\overrightarrow{\mathbf{h}_T},\overleftarrow{\mathbf{h}_T}]$, where $\overrightarrow{\mathbf{h}_T}$ and $\overleftarrow{\mathbf{h}_T}$ are the encoded vectors summarizing the past and the future, respectively.

\subsubsection{Results} 

In our experiments, we use the neural architecture shown in Figure \ref{fig:architecture} with one notable difference: we exclude the averaged semantic vectors (\wiki), and instead we use \new{either a CNN (Figure \ref{fig:cnn}) or an LSTM-RNN (Figure~\ref{fig:lstm-framework})} to encode the vectors for the sentences (\new{i.e., one reference and two candidate translations}). \new{The objective function remains the same as in Equation \ref{eq:logcost}.} 

Complex neural models like LSTMs tend to overfit because of the increased number of parameters. In order to avoid overfitting, we use dropout \cite{hinton:dropout} of embedding and hidden units and we perform early stopping based on the accuracy on the development set. We experimented with the following dropout rates: $\{0.0, 0.25, 0.5, 0.75\}$. To compare to the best baseline results, we initialize $E$ with the pretrained \wiki\ word vectors \cite{pennington-socher-manning:2014:EMNLP2014}, and we do fine-tuning of these vectors. \new{For CNN, we experimented with $\{50, 100, 150\}$ number of filters, and we use filtering and pooling lengths of $\{3, 4, 5\}$. For LSTM,} we experimented with $\{50, 100, 150\}$ number of hidden units in the LSTM layer. \new{These parameters are optimized on the development set.}

Table \ref{tab:lstm-discussion} shows the results of our models on the WMT12 testset. \new{The first row shows the results for the averaging baseline (no semantic composition using CNN or LSTM). The second and the third rows show the results for our CNN model: when it is initialized with pretrained \wiki\ vectors, and when it is randomly initialized, respectively. We can see that CNN with pre-trained vectors is slightly better than our averaging BOW baseline. } 

\begin{table*}[t]
\centering
{\footnotesize\begin{tabular}{lccccc}
\toprule
   & \multicolumn{5}{c}{\bf Kendall's $\tau$} \\\cmidrule(l{2pt}r{2pt}){2-6}
\multicolumn{1}{c}{\bf Details} & { \bf cz }&{\bf de} & {\bf es}  & {\bf fr} &{\bf AVG}\\
\midrule
Averaging (no LSTM) & 26.92  & 33.05 &  29.83 &  29.01&  29.70\\
\midrule
\new{CNN}         & 26.47  & 32.28 & 30.24  & 28.96 &  29.77\\
\new{CNN(random)} & 25.80   & 33.61 & 29.62  & 28.83 & 29.55 \\
\midrule
Unidirectional LSTM  & 25.51 & 33.31 & 30.40 & 29.16 & 29.59\\
Bidirectional LSTM   & 26.20 & 33.82 & 30.16 & 29.23 & \textbf{29.85} \\
Bidirectional LSTM  (random) & 25.91 & 33.81 & 30.44 & 28.90 &  29.76 \\
\bottomrule
\end{tabular}}
\caption{\label{tab:lstm-discussion}{\small Kendall's $\tau$ on WMT12 for different variants of LSTM-RNNs. }}
\end{table*}

The \new{fourth} row shows the results of our model with a unidirectional LSTM. We can notice that even though the unidirectional LSTM outperforms the baseline in three out of four languages, it fails to beat the baseline on average because of its poor performance on Czech. 
%
Bidirectional LSTM (fifth row) yields an average improvement of $+0.15$ over the baseline. Finally, the sixth row shows the results of the model when word vectors are randomly initialized (as oppposed to pretrained). 
This model performs slightly better than the baseline on average, which means that bidirectional LSTMs can achieve good results even without pretrained word vectors. 

We notice that our models with LSTM-based semantic composition fail to achieve better results for Czech. This could be due to the reordering errors made by Czech-English traslation systems, for which sequential LSTMs  may not be robust enough to encode the necessary information. 

Another general observation is that fine tuning and composition with LSTMs did not yield improvements to the extent that we had expected. One potential reason could be that the compositional aspect is partially captured by the syntactic embeddings, which are produced as the parser composes phrases hierarchically using a recursive neural network \cite{socher-EtAl:2013:ACL2013}. In order to investigate this, 
Table~\ref{tab:lstm-discussion2} shows the results of the baseline and the model with LSTM after excluding the syntactic embeddings. In this setting, LSTM yields a larger gain of $+0.31$.


\begin{table*}[t]
\centering
{\footnotesize\begin{tabular}{@{}l@{ }@{ }ccccc@{}}
\toprule
   & \multicolumn{5}{c}{\bf Kendall's $\tau$} \\\cmidrule(l{2pt}r{2pt}){2-6}
\multicolumn{1}{c}{\bf Details} & { \bf cz }&{\bf de} & {\bf es}  & {\bf fr} &{\bf AVG}\\
\midrule
Averaging baseline, without syntactic embeddings & 25.67 & 32.50 & 29.21 & {28.92} & 29.07 \\
Bidirectional LSTM, without syntactic embeddings  & 26.31 & 32.60 & 29.31 & 29.20 &  29.38 \\
\bottomrule
\end{tabular}}
\caption{\label{tab:lstm-discussion2}{\small Results for baseline and LSTM-based model without syntactic embeddings. }}
\end{table*}

%% file: absolute-scores.tex

\noindent In this section, we show how we can use the pairwise NN architecture to produce absolute quality scores when the input is reduced to a single translation, i.e., we turn our pairwise metric into a standard metric for MT evaluation. We further compare the quality of this metric with the state of the art on two WMT datasets,
both at the sentence and at the system levels.

\subsection{Generating an Absolute Score}

\noindent As we have a pairwise MT evaluation approach, in our experiments above,
we always compared two translations.
While arguably, this is a setup that is useful in many situations, 
most MT evaluation metrics are designed to assign absolute scores
for the output of a single system.
Below we show how we can turn our pairwise metric into
such a metric.

In order to generate an absolute score for a translation $t$ of a particular sentence from a particular system, without the need to use the translations of other systems, we provide to our neural network the vectors for that translation paired with an \emph{empty} translation vector $\emptytrans$; we also provide the vector for the reference as normal. We handle the pairwise features in a similar fashion, using empty values.
We experiment with two simple strategies to generate empty vectors and values:
\begin{itemize}
\item[(a)] \emph{using zeroes}, and 
\item[(b)] \emph{using average values} for each vector coordinate or pairwise feature, 
averaging over the examples seen in the training input.
\end{itemize}


In either case, we ask the NN for two predictions, one using empty values for translation $t_1$, and another one with empty values for translation $t_2$, i.e., we plug the single translation $t$ vector as $t_1$ with empty values for $t_2$ to obtain a prediction $p(t,\emptytrans,r)$, and once as $t_2$ with empty values for $t_1$, which yields a prediction $p(\emptytrans,t,r)$. We then subtract the scores for the two predictions to generate the final score for the sentence: $p(t,\emptytrans,r) - p(\emptytrans,t,r)$.

Note that we do not use just one of the two predictions,\footnote{Using $p(t,\emptytrans,r)$ or $p(\emptytrans,t,r)$, instead of their difference, yielded slightly lower results.} $p(t,\emptytrans,r)$ or $p(\emptytrans,t,r)$, as our network is not exactly symmetric. 
By subtracting the two, i.e.,
the score for $t$ winning over an average translation and 
the score for $t$ losing to an average translation,
we look at the margin between winning vs. losing to an average translation.

Note that our technique is similar to that used by PRO for tuning machine translation parameters \cite{Hopkins2011},
where training is done in a pairwise fashion by subtracting the vectors for the two competing translations and then training to predict +1 or -1.
At test time, a vector for a single translation is used, which is equivalent to subtracting a zero vector from it, i.e., to predicting whether the translation would win against an empty 
translation, and by what margin.

The top three lines of Table~\ref{t:comparison-sentence-level:WMT2012} show a comparison of the absolute vs. the pairwise version of our neural-based metric. We refer to our metric as NNRK (for Neural Network ReranKing), using subindices to describe the metric variant. The comparison is on the WMT12 dataset, at the segment level. We can see that using absolute scores instead of pairwise comparisons yields 
better results: by 0.9-1.2 Kendall's $\tau$ points absolute.
We believe that this is because by using an absolute score rather than a pairwise decision, we remove some possible circularities, e.g., in the pairwise framework, we could predict that translation $x$ is better than $y$, and $y$ is better than $z$, but $z$ is better than $x$. This is not possible when working with absolute scores.

We further see that comparing to an average vector is slightly better than comparing to a zero one, but the difference is not large: 0.3 Kendall's $\tau$ points absolute.\footnote{We normalize the input to the NN to the $[-1,1]$ interval, and we further train our NN in a symmetric way, where each pair of translations is used twice: once as a positive, and once as a negative example. As a result, the average value for each vector coordinate or for each pairwise feature is close to zero, and thus, the two approaches yield very similar results.}

\begin{table*}[t]
\centering
{\footnotesize\begin{tabular}{l@{\hspace*{0.25cm}}l@{\hspace*{0.25cm}}c@{\hspace*{0.15cm}}c@{\hspace*{0.15cm}}c@{\hspace*{0.15cm}}c@{\hspace*{0.1cm}}c}
\toprule
&& \multicolumn{5}{c}{\bf Kendall's $\tau$ {\scriptsize (WMT12-style)}} \\
{\bf System}& {\bf Details} & { \bf cz }&{\bf de} & {\bf es} & {\bf fr} &{\bf AVG}\\
\midrule
NNRK$_{\rm Mean}$ & multi-layer NN, mean vector & \bf 27.7 & \bf 34.7 & \bf 31.4 & \bf 29.7 & \bf 30.9 \\
NNRK$_{\rm Zero}$ & multi-layer NN, zero vector & 27.3 & 34.5 & \bf 31.4 & 29.2 & 30.6 \\
NNRK$_{\rm pairwise}$ & multi-layer NN, pairwise & 26.3 & 33.2 & 30.4 & 28.9 & 29.7\\
\midrule
\disco~\cite{discoMT:WMT2014} & Best on the WMT12 dataset & \emph{na} & \emph{na} & \emph{na} & \emph{na} &  30.2 \\
\spede~\cite{Wang2012} & 1st at the WMT12 competition & 21.2 & 27.8 & 26.5 & 26.0 & 25.4\\
\meteor~\cite{Denkowski2011}   & 2nd at the WMT12 competition  & 21.2 & 27.5 & 24.9 & 25.1 & 24.7 \\
Guzm\'an et al. \cite{guzman-EtAl:2014} & Preference kernel approach & 23.1 & 25.8 & 22.6 & 23.2 & 23.7\\
\amber~\cite{chen-kuhn-foster:2012:WMT}    & 3rd at the WMT12 competition  & 19.1 & 24.8 & 23.1 & 24.5 & 22.9 \\
\bottomrule 
\end{tabular}}
\caption{\label{t:comparison-sentence-level:WMT2012}{\small Comparing to the state of the art at the segment level on the WMT12 dataset, translating into English. Values marked as \emph{na} were not reported by the authors.}}
\end{table*}

\subsection{Comparison to the State of the Art}

\noindent Below we compare the performance of our NNRK metric to the state of the art on WMT12 and WMT14, 
both at the segment and at the system level.

\paragraph{WMT12, segment-level} 
Table~\ref{t:comparison-sentence-level:WMT2012} shows that our NNRK metric with absolute scores outperforms the best previously published results on the WMT12 dataset, at the segment level: both overall, and for each of the four individual language pairs. It is over five Kendall's $\tau$ points absolute better than the best system at the WMT12 competition (we show the top-3 systems from WMT12 to put the results in perspective). It also outperforms, by 0.7 points absolute, the tuned \disco\ metric~\cite{discoMT:WMT2014}, which had achieved the best results on the WMT12 dataset. Moreover, it outperforms by a margin our preference kernel approach \cite{guzman-EtAl:2014}.

\begin{table*}[t]
\centering
{\footnotesize\begin{tabular}{l@{\hspace*{0.25cm}}l@{\hspace*{0.25cm}}c@{\hspace*{0.2cm}}c@{\hspace*{0.2cm}}c@{\hspace*{0.2cm}}c@{}c}
\toprule
&& \multicolumn{5}{c}{\bf Spearman's rank correlation $\rho$} \\
{\bf System}& {\bf Details} & { \bf cz }&{\bf de} & {\bf es} & {\bf fr} &{\bf AVG}\\
\midrule
NNRK$_{\rm Mean}$ & multi-layer NN, mean vector & \bf 94.3 & \bf 92.9 & \bf 97.9 & \bf 86.4 & \bf 92.9 \\
NNRK$_{\rm Zero}$ & multi-layer NN, zero vector & \bf 94.3 & 92.1 & 96.5 & 85.4 & 92.1 \\
\midrule
\disco~\cite{discoMT:WMT2014} & Best on the WMT12 dataset & \emph{na} & \emph{na} & \emph{na} & \emph{na} &  91.5 \\
\sc SEMPOS~\cite{machavcek-bojar:2011:WMT} & 1st at the WMT12 competition & \bf 94.3 & 92.4 & 93.7 & 80.4 & 90.2\\
\amber~\cite{chen-kuhn-foster:2012:WMT}  & 2nd at the WMT12 competition & 82.9 & 78.5 & 96.5 & 85.0 & 85.7\\
\meteor~\cite{Denkowski2011} & 3rd at the WMT12 competition & 65.7 & 88.5 & 95.1 & 84.3 & 83.4\\
\bottomrule 
\end{tabular}}
\caption{\label{t:comparison-system-level:WMT2012}{\small Comparing NNRK to the state of the art at the system level on the WMT12 dataset, translating into English.}}
\end{table*}

\paragraph{WMT12, system-level} 
We further converted our segment-level scores to system-level ones, to produce a system-level version of our NNRK metric. For this purpose, we first calculated a score for each test sentence, and then we took the average of these scores.\footnote{We also tried an aggregation based on the sign of these scores, i.e., by the number of wins over the average translation, but it worked a bit worse.}

Following, the evaluation setup of WMT12, we used Spearman's rank correlation to compare our system-level scores to those assigned by human judges.\footnote{See \cite{WMT12} for a discussion about how the human pairwise segment-level judgments are aggregated to produce a human system-level score.}
To calculate Spearman's rank correlation, we first convert the raw scores assigned to each system to ranks, and then we use the following formula \cite{Spearman}:
\begin{equation}
\rho = 1 - \frac{6 \sum{d_{i}^2}}{n(n^2-1)}
\end{equation}
\noindent where $d_i$ is the difference between the ranks for system $i$, and $n$ is the number of systems being evaluated.

Note that this formula requires that there be no ties in the ranks of the systems (based on the automatic metric or based on the human judgments), which was indeed the case. Spearman's rank correlation ranges between -1 and +1. 
In our experiments, we used the official script from WMT12 for the score calculation, in order to ensure direct comparability of our results to those from the WMT12 shared task.

The results are shown in Table~\ref{t:comparison-system-level:WMT2012}.
Not surprisingly, they are on par with what we saw at the segment level.
Once again, our NNRK metric outperformed the metrics from the WMT12 competition, as well as the best post-competition result of \disco; moreover, our NNRK metric is strong across all language pairs.
Also, as expected, the system-level score based on the NNRK$_{\rm Mean}$ segment-level score performed better than the one based on NNRK$_{\rm Zero}$.

\paragraph{WMT14, segment-level} 
We further compared our results to those on the WMT14 metrics task. This allows us to evaluate and compare our metric on another dataset, and also to compare directly to {\sc ReVal} \cite{gupta-orasan-vangenabith:2015:EMNLP}, the evaluation metric based on recurrent neural networks discussed in Section~\ref{sec:related}.
Note that we did not retrain or tune our NNRK metric on the WMT12 or WMT13 data that was available by the time of the WMT14 competition. We simply apply the same network trained on WMT11 to the test set from WMT14

The results are shown in Table~\ref{t:comparison-sentence-level:WMT2014}.
Once again, our NNRK metric is very competitive and outperforms all rivals, except for the tuned \disco, which combines about twenty strong pre-existing evaluation metrics (in addition to discourse-based kernels), while we only incorporate four pre-existing metrics.\footnote{\new{The high-complexity of the best scoring metrics was also observed in WMT 2015 and WMT 2016. The best performing metric at the segment level was {\sc DPMFcomb}~\cite{yu-etal:2015:WMT} a syntactic metric (DPMF) combined with a massive number of preexisting metrics provided by \asiya. Comparatively, our NNRK proposal is much simpler.}} Note that we outperform 
the {\sc ReVal} NN metric by more than six Kendall's $\tau$ points absolute; we are also better on all four language pairs. Overall, we are best on Hindi-English, but we are surprisingly weak on Russian-English.

\begin{table*}[t]
\centering
\hspace*{-3mm}
{\footnotesize\begin{tabular}{l@{\hspace*{0.15cm}}l@{\hspace*{0.15cm}}c@{\hspace*{0.15cm}}c@{\hspace*{0.15cm}}c@{\hspace*{0.15cm}}c@{\hspace*{0.15cm}}c@{\hspace*{0.15cm}}c}
\toprule
&& \multicolumn{6}{c}{\bf Kendall's $\tau$ {\scriptsize (WMT14-style)}} \\
{\bf System}& {\bf Details} & { \bf fr }&{\bf de} & {\bf hi} & {\bf cs} & {\bf ru} &{\bf AVG}\\
\midrule
NNRK$_{\rm Mean}$ & multi-layer NN, mean vector & 41.3 & 36.5 & \bf 44.1 & 31.8 & 30.2 & 36.8 \\
NNRK$_{\rm Zero}$ & multi-layer NN, zero vector & 41.0 & 36.9 & 43.7 & 31.8 & 30.0 & 36.7 \\
\midrule
\disco~\cite{discoMT:WMT2014} & 1st at the WMT14 competition & \bf 43.3 & \bf 38.0 & 43.4 & \bf 32.8 & \bf 35.5 & \bf 38.6 \\
\sc BEER \cite{stanojevic-simaan:2014:W14-33} & 2nd at the WMT14 competition & 41.7 & 33.7 & 43.8 & 28.4 & 33.3 & 36.2\\
\sc REDcombSent \cite{wu-yu-liu:2014:W14-33} & 3rd at the WMT14 competition & 40.6 & 33.8 & 41.7 & 28.4 & 33.6 & 35.6\\
\sc REDcombSysSent \cite{wu-yu-liu:2014:W14-33} & 3rd at the WMT14 competition & 40.8 & 33.8 & 41.6 & 28.2 & 33.6 & 35.6\\
\sc ReVal \cite{gupta-orasan-vangenabith:2015:EMNLP} & RNN-based evaluation measure & 34.7 & 27.9 & 36.7 & 25.2 & 27.4 & 30.4\\
\bottomrule 
\end{tabular}}
\caption{\label{t:comparison-sentence-level:WMT2014}{\small Comparing to the state of the art at the segment level on the WMT14 dataset, translating into English.}}
\end{table*}

Note that the version of Kendall's $\tau$ reported here, which is the one used for official evaluation at WMT14, is slightly different from that used in WMT12 due due to different handling of ties. See the WMT14 Metrics Task overview paper \cite{machacek-bojar:2014:W14-33}, which discusses the issue in detail.
The numbers we report here are calculated using the official scorer from WMT14.

\paragraph{WMT14, system-level} 
The WMT14 results at the system level are shown in Table~\ref{t:comparison-system-level:WMT2014}. Our NNRK metric is the third in the ranking, after \disco\ and {\sc ReVal}. While \disco\ is best overall, it is not the strongest on any of the five languages in the table. In contrast, NNRK has the highest scores on French-English and German-English, while {\sc ReVal} is strongest on Czech-English and Russian-English, and {\sc LAYERED} is best on Hindi-English. On this dataset, NNRK is slightly below {\sc ReVal}; this can be expected, as our metric specializes on segment-level pairwise judgments and it was trained with only the smaller dataset from WMT11. Note that again our overall correlation is strongly penalized by a very low score on Russian-English. The reason for this phenomenon has to be further investigated.

Note that this time we used \emph{Pearson correlation} \cite{Pearson}, as it was the official system-level score at WMT14. This is a more general correlation coefficient than Spearman's and does not require that all $n$ ranks be distinct integers. It ranges between -1 and +1, where higher absolute score is better. We used the official WMT14 scoring scripts to calculate it.  

%

\begin{table*}[t]
\centering
\hspace*{-3mm}
{\footnotesize\begin{tabular}{l@{\hspace*{0.15cm}}l@{\hspace*{0.15cm}}c@{\hspace*{0.15cm}}c@{\hspace*{0.15cm}}c@{\hspace*{0.15cm}}c@{\hspace*{0.15cm}}c@{\hspace*{0.15cm}}c}
\toprule
&& \multicolumn{6}{c}{\bf Pearson correlation $r$} \\
{\bf System}& {\bf Details} & { \bf fr }&{\bf de} & {\bf hi} & {\bf cs} & {\bf ru} &{\bf AVG}\\
\midrule
NNRK$_{\rm Mean}$ & multi-layer NN, mean vector & \bf 98.4 & \bf 95.4 & 96.6 & 98.2 & 76.7 & 93.0\\
NNRK$_{\rm Zero}$ & multi-layer NN, zero vector & 97.9 & 93.2 & 95.1 & 97.3 & 76.0 & 91.9\\
\midrule
\disco~\cite{discoMT:WMT2014} & 1st at the WMT14 competition & 97.7 & 94.3 & 95.6 & 97.5 & 87.0 & \bf 94.4 \\
\sc ReVal \cite{gupta-orasan-vangenabith:2015:EMNLP} & RNN-based evaluation measure & 97.9 & 90.6 & 91.8 & \bf 99.4 & \bf 88.1 & 93.5\\
\sc LAYERED \cite{gautam-bhattacharyya:2014:W14-33} & 2nd at the WMT14 competition & 97.3 & 89.3 & \bf 97.6 & 94.1 & 85.4 & 92.7\\
\sc \disco$_{\hbox{\footnotesize{\sc untuned}}}$~\cite{discoMT:WMT2014} & 3nd at the WMT14 competition & 97.0 & 92.1 & 86.2 & 98.3 & 85.6 & 91.8\\
\bottomrule 
\end{tabular}}
\caption{\label{t:comparison-system-level:WMT2014}{\small Comparing to the state of the art at the system level on the WMT14 dataset, translating into English.}}
\end{table*}

%% file: conclusions.tex


\noindent We have presented a framework for learning a tunable MT evaluation metric
which operates in a pairwise ranking setting, and is trained on pre-existing pairwise human preference judgments.


As our basic model, we used a feed-forward neural network, which is trained to differentiate better from worse translations. The input layer encodes lexical, syntactic and semantic information from the reference and from the two translation hypotheses, which are efficiently compacted into relatively small embeddings. The network has a hidden layer, motivated by our intuition about the problem, which captures the interactions among the relevant input components. It is also able to incorporate the prediction from external pre-existing MT evaluation measures as direct features to the output layer. 


Unlike previously proposed kernel-based approaches~\cite{guzman-EtAl:2014} or heavy combination-based metrics like \disco~\cite{discoMT:WMT2014}, our framework allows us to do both training and inference efficiently.
Results when evaluating in a pairwise setup have shown that our basic NN model yields state-of-the-art results when using lexical, syntactic and semantic features in combination with four standard MT evaluation metrics.
Moreover, we have shown that the contribution of the different information sources is additive, demonstrating that the framework can effectively integrate complementary information. Also, we have presented evidence showing that using the hidden layers is advantageous over a linear pairwise classification model.


We have investigated several extensions over the basic model. First, we have demonstrated that the neural network can be trained to optimize a task-specific cost function, which is more appropriate for the pairwise MT evaluation setting.
Second, we have played with different granularities of features, such as $n$-gram matches and other components of \bleu, which individually work better than using the aggregated \bleu\ score. Third, we have explored the possibility of fine-tuning the embeddings with feedback from the task. Finally, we have implemented a semantic representation of the input sentences by using convolutional and recurrent neural networks (concretely, CNNs and bidirectional LSTMs). The last two extensions have shown that some little improvements in performance are attainable at the cost of increased complexity and lower efficiency. The trade-off is to be resolved in terms of practical needs.


Finally, we have shown that we can use the network trained pairwise to produce absolute translation quality scores for single translations. The main idea is to estimate whether the translation would win or lose against an \emph{empty} average translation, and by what margin. The derived NNRK metric performs comparably to the state of the art on benchmark WMT datasets at the system level, but particularly at the segment level. NNRK yields the absolute best results on the WMT12 test dataset. For WMT14, the results are also good, not far from the best ones published (i.e., those of the heavy combination metric \disco). Compared to the other existing neural evaluation metrics, ReVal, NNRK performs significantly better at the segment level, and it is comparable at the system level.


In future work, we plan to study other aspects to complement the present work, including, among others: \Ni the differences and stability of the metric across language pairs, especially in the light of the surprisingly low results obtained for the Russian-English language pair, and \Nii the robustness of the NNRK metric, including the relevance of the training set, the impact of the quality of the syntactic analysis on the hardly grammatical translations, etc.

Finally, we would also like to incorporate features from the \emph{source} sentence. We believe that our framework can support learning similarities between the two translations and the source, for an improved MT evaluation. Variations of this architecture might be useful for related tasks such as Quality Estimation and hypothesis re-ranking for Machine Translation, where no references are available. Searching for a lightweight, fast evaluation metric configuration that can be used for extensive MT evaluation with the best possible correlation with human annotations is also one of our objectives for the near future.

%% file: neural-mte-csl.bbl
\begin{thebibliography}{10}
\expandafter\ifx\csname url\endcsname\relax
  \def\url#1{\texttt{#1}}\fi
\expandafter\ifx\csname urlprefix\endcsname\relax\def\urlprefix{URL }\fi
\expandafter\ifx\csname href\endcsname\relax
  \def\href#1#2{#2} \def\path#1{#1}\fi

\bibitem{Papineni:Roukos:Ward:Zhu:2002}
K.~Papineni, S.~Roukos, T.~Ward, W.-J. Zhu, {BLEU}: a method for automatic
  evaluation of machine translation, in: Proceedings of 40th Annual Meting of
  the Association for Computational Linguistics, ACL~'02, Philadelphia,
  Pennsylvania, USA, 2002, pp. 311--318.

\bibitem{Lavie:2009:MMA}
A.~Lavie, M.~Denkowski, The {METEOR} metric for automatic evaluation of machine
  translation, Machine Translation 23~(2--3) (2009) 105--115.

\bibitem{Gimenez2007}
J.~Gim\'{e}nez, L.~M\`{a}rquez, Linguistic features for automatic evaluation of
  heterogenous {MT} systems, in: Proceedings of the Second Workshop on
  Statistical Machine Translation, WMT '07, Prague, Czech Republic, 2007, pp.
  256--264.

\bibitem{Popovic2007}
M.~Popovi\'{c}, H.~Ney, Word error rates: Decomposition over {POS} classes and
  applications for error analysis, in: Proceedings of the Second Workshop on
  Statistical Machine Translation, WMT '07, Prague, Czech Republic, 2007, pp.
  48--55.

\bibitem{Liu2005}
D.~Liu, D.~Gildea, Syntactic features for evaluation of machine translation,
  in: Proceedings of the ACL Workshop on Intrinsic and Extrinsic Evaluation
  Measures for Machine Translation and/or Summarization, Ann Arbor, Michigan,
  USA, 2005, pp. 25--32.

\bibitem{Lo2012}
C.-k. Lo, A.~K. Tumuluru, D.~Wu, Fully automatic semantic {MT} evaluation, in:
  Proceedings of the Seventh Workshop on Statistical Machine Translation, WMT
  '12, Montr{\'e}al, Canada, 2012, pp. 243--252.

\bibitem{Comelles2010}
E.~Comelles, J.~Gim\'enez, L.~M\`arquez, I.~Castell\'on, V.~Arranz,
  Document-level automatic {MT} evaluation based on discourse representations,
  in: Proceedings of the Joint Fifth Workshop on Statistical Machine
  Translation and MetricsMATR, WMT '10, Uppsala, Sweden, 2010, pp. 333--338.

\bibitem{Wong2012}
B.~Wong, C.~Kit, Extending machine translation evaluation metrics with lexical
  cohesion to document level, in: Proceedings of the 2012 Joint Conference on
  Empirical Methods in Natural Language Processing and Computational Natural
  Language Learning, EMNLP-CoNLL '12, Jeju Island, Korea, 2012, pp. 1060--1068.

\bibitem{discoMT:acl2014}
F.~Guzm\'{a}n, S.~Joty, L.~M\`{a}rquez, P.~Nakov, Using discourse structure
  improves machine translation evaluation, in: Proceedings of 52nd Annual
  Meeting of the Association for Computational Linguistics, ACL '14, Baltimore,
  Maryland, USA, 2014, pp. 687--698.

\bibitem{discoMT:WMT2014}
S.~Joty, F.~Guzm\'{a}n, L.~M\`{a}rquez, P.~Nakov, Disco{TK}: Using discourse
  structure for machine translation evaluation, in: Proceedings of the Ninth
  Workshop on Statistical Machine Translation, WMT '14, Baltimore, Maryland,
  USA, 2014, pp. 402--408.

\bibitem{WMT-MT14}
M.~Machacek, O.~Bojar, Proceedings of the Ninth Workshop on Statistical Machine
  Translation, Association for Computational Linguistics, 2014, Ch. Results of
  the WMT14 Metrics Shared Task, pp. 293--301.

\bibitem{WMT-MT15}
M.~Stanojevi\'{c}, A.~Kamran, P.~Koehn, O.~Bojar, Results of the wmt15 metrics
  shared task, in: Proceedings of the Tenth Workshop on Statistical Machine
  Translation, Association for Computational Linguistics, Lisbon, Portugal,
  2015, pp. 256--273.

\bibitem{callisonburch-EtAl:2007:WMT}
C.~Callison-Burch, C.~Fordyce, P.~Koehn, C.~Monz, J.~Schroeder, ({M}eta-)
  evaluation of machine translation, in: Proceedings of the Second Workshop on
  Statistical Machine Translation, WMT '07, Prague, Czech Republic, 2007, pp.
  136--158.

\bibitem{albrecht:2008}
J.~Albrecht, R.~Hwa, Regression for machine translation evaluation at the
  sentence level, Machine Translation 22~(1-2) (2008) 1--27.

\bibitem{Duh:2008}
K.~Duh, Ranking vs. regression in machine translation evaluation, in:
  Proceedings of the Third Workshop on Statistical Machine Translation,
  WMT~'08, Columbus, Ohio, USA, 2008, pp. 191--194.

\bibitem{song-cohn:2011:WMT}
X.~Song, T.~Cohn, Regression and ranking based optimisation for sentence-level
  {MT} evaluation, in: Proceedings of the Sixth Workshop on Statistical Machine
  Translation, WMT '11, Edinburgh, Scotland, 2011, pp. 123--129.

\bibitem{guzman-EtAl:2014}
F.~Guzm\'{a}n, S.~Joty, L.~M\`{a}rquez, A.~Moschitti, P.~Nakov, M.~Nicosia,
  Learning to differentiate better from worse translations, in: Proceedings of
  the 2014 Conference on Empirical Methods in Natural Language Processing,
  EMNLP~'14, Doha, Qatar, 2014, pp. 214--220.

\bibitem{guzman-EtAl:2015:ACL-IJCNLP}
F.~Guzm\'{a}n, S.~Joty, L.~M\`{a}rquez, P.~Nakov, Pairwise neural machine
  translation evaluation, in: Proceedings of the 53rd Annual Meeting of the
  Association for Computational Linguistics and the 7th International Joint
  Conference on Natural Language Processing (Volume 1: Long Papers),
  Association for Computational Linguistics, Beijing, China, 2015, pp.
  805--814.

\bibitem{WMT12}
C.~Callison-Burch, P.~Koehn, C.~Monz, M.~Post, R.~Soricut, L.~Specia, Findings
  of the 2012 {W}orkshop on {S}tatistical {M}achine {T}ranslation, in:
  Proceedings of the Seventh Workshop on Statistical Machine Translation,
  WMT~'12, Montr{\'e}al, Canada, 2012, pp. 10--51.

\bibitem{stanojevic-simaan:2015:WMT}
M.~Stanojevi\'{c}, K.~Sima'an, Beer 1.1: Illc uva submission to metrics and
  tuning task, in: Proceedings of the Tenth Workshop on Statistical Machine
  Translation, Association for Computational Linguistics, Lisbon, Portugal,
  2015, pp. 396--401.

\bibitem{pado-EtAl:2009:ACLIJCNLP}
S.~Pado, M.~Galley, D.~Jurafsky, C.~D. Manning, Robust machine translation
  evaluation with entailment features, in: Proceedings of the Joint Conference
  of the 47th Annual Meeting of the ACL and the 4th International Joint
  Conference on Natural Language Processing of the AFNLP, Association for
  Computational Linguistics, Suntec, Singapore, 2009, pp. 297--305.

\bibitem{kulesza2004}
A.~Kulesza, S.~M. Shieber, A learning approach to improving sentence-level {MT}
  evaluation, in: Proceedings of the 10th International Conference on
  Theoretical and Methodological Issues in Machine Translation, 2004.

\bibitem{Bengio03}
Y.~Bengio, R.~Ducharme, P.~Vincent, C.~Janvin, A neural probabilistic language
  model, Journal of Machine Learning Research 3 (2003) 1137--1155.

\bibitem{Mikolov10}
T.~Mikolov, M.~Karafi{\'{a}}t, L.~Burget, J.~Cernock{\'{y}}, S.~Khudanpur,
  Recurrent neural network based language model, in: 11th Annual Conference of
  the International Speech Communication Association, Makuhari, Chiba, Japan,
  2010, pp. 1045--1048.

\bibitem{Devlin14}
J.~Devlin, R.~Zbib, Z.~Huang, T.~Lamar, R.~Schwartz, J.~Makhoul, Fast and
  robust neural network joint models for statistical machine translation, in:
  Proceedings of the 52nd Annual Meeting of the Association for Computational
  Linguistics, ACL~'14, Baltimore, Maryland, USA, 2014, pp. 1370--1380.

\bibitem{kalchbrenner13}
N.~Kalchbrenner, P.~Blunsom, Recurrent continuous translation models, in:
  Proceedings of the 2013 Conference on Empirical Methods in Natural Language
  Processing, Seattle, Washington, USA, 2013, pp. 1700--1709.

\bibitem{SutskeverVL14}
I.~Sutskever, O.~Vinyals, Q.~V. Le, Sequence to sequence learning with neural
  networks, in: Proceedings of the Neural Information Processing Systems,
  NIPS~'14, Montreal, Canada, 2014.

\bibitem{chen-etal:2015:WMT}
B.~Chen, H.~Guo, R.~Kuhn, Multi-level evaluation for machine translation, in:
  Proceedings of the Tenth Workshop on Statistical Machine Translation,
  Association for Computational Linguistics, Lisbon, Portugal, 2015, pp.
  361--365.

\bibitem{chen-guo:2015:ACL-IJCNLP}
B.~Chen, H.~Guo, \href{http://www.aclweb.org/anthology/P15-2025}{Representation
  based translation evaluation metrics}, in: Proceedings of the 53rd Annual
  Meeting of the Association for Computational Linguistics and the 7th
  International Joint Conference on Natural Language Processing (Volume 2:
  Short Papers), Association for Computational Linguistics, Beijing, China,
  2015, pp. 150--155.
\newline\urlprefix\url{http://www.aclweb.org/anthology/P15-2025}

\bibitem{gupta-orasan-vangenabith:2015:WMT}
R.~Gupta, C.~Orasan, J.~van Genabith, Machine translation evaluation using
  recurrent neural networks, in: Proceedings of the Tenth Workshop on
  Statistical Machine Translation, Association for Computational Linguistics,
  Lisbon, Portugal, 2015, pp. 380--384.

\bibitem{gupta-orasan-vangenabith:2015:EMNLP}
R.~Gupta, C.~Orasan, J.~van Genabith, {ReVal}: A simple and effective machine
  translation evaluation metric based on recurrent neural networks, in:
  Proceedings of the 2015 Conference on Empirical Methods in Natural Language
  Processing, Lisbon, Portugal, 2015, pp. 1066--1072.

\bibitem{bojar-EtAl:2016:WMT2}
O.~Bojar, Y.~Graham, A.~Kamran, M.~Stanojevi\'{c}, Results of the wmt16 metrics
  shared task, in: Proceedings of the First Conference on Machine Translation,
  Berlin, Germany, 2016, pp. 199--231.

\bibitem{guzman-EtAl:2016:ACL}
F.~Guzm\'{a}n, L.~M\`{a}rquez, P.~Nakov, Machine translation evaluation meets
  community question answering, in: \hbox{To appear in} Proceedings of the 54th
  Annual Meeting of the Association for Computational Linguistics (ACL 2016),
  Association for Computational Linguistics, Berlin, Germany, 2016.

\bibitem{Duchi11}
J.~Duchi, E.~Hazan, Y.~Singer, Adaptive subgradient methods for online learning
  and stochastic optimization, Journal of Machine Learning Research 12 (2011)
  2121--2159.

\bibitem{bergstra+al:2010-scipy}
J.~Bergstra, O.~Breuleux, F.~Bastien, P.~Lamblin, R.~Pascanu, G.~Desjardins,
  J.~Turian, D.~Warde-Farley, Y.~Bengio, Theano: a {CPU} and {GPU} math
  expression compiler, in: Proceedings of the Python for Scientific Computing
  Conference, SciPy~'10, Austin, Texas, 2010.

\bibitem{socher-EtAl:2013:ACL2013}
R.~Socher, J.~Bauer, C.~D. Manning, N.~Andrew~Y., Parsing with compositional
  vector grammars, in: Proceedings of the 51st Annual Meeting of the
  Association for Computational Linguistics (Volume 1: Long Papers), ACL~'13,
  Sofia, Bulgaria, 2013, pp. 455--465.

\bibitem{Mitchell:Lapata:2010}
J.~Mitchell, M.~Lapata, Composition in distributional models of semantics,
  Cognitive Science 34~(8) (2010) 1388--1439.

\bibitem{pennington-socher-manning:2014:EMNLP2014}
J.~Pennington, R.~Socher, C.~Manning, Glove: Global vectors for word
  representation, in: Proceedings of the Conference on Empirical Methods in
  Natural Language Processing, EMNLP~'14, Doha, Qatar, 2014, pp. 1532--1543.

\bibitem{P14-1023}
M.~Baroni, G.~Dinu, G.~Kruszewski, Don't count, predict! {A} systematic
  comparison of context-counting vs. context-predicting semantic vectors, in:
  Proceedings of the 52nd Annual Meeting of the Association for Computational
  Linguistics, ACL~'14, Baltimore, Maryland, USA, 2014, pp. 238--247.

\bibitem{mikolov-yih-zweig:2013:NAACL-HLT}
T.~Mikolov, W.-t. Yih, G.~Zweig, Linguistic regularities in continuous space
  word representations, in: Proceedings of the 2013 Conference of the North
  American Chapter of the Association for Computational Linguistics: Human
  Language Technologies, NAACL-HLT~'13, Atlanta, Georgia, USA, 2013, pp.
  746--751.

\bibitem{WMT11}
C.~Callison-Burch, P.~Koehn, C.~Monz, O.~Zaidan, Findings of the 2011 workshop
  on statistical machine translation, in: Proceedings of the Sixth Workshop on
  Statistical Machine Translation, WMT '11, Edinburgh, Scotland, 2011, pp.
  22--64.

\bibitem{WMT13}
M.~Mach\'{a}\v{c}ek, O.~Bojar, Results of the {WMT13} metrics shared task, in:
  Proceedings of the Eighth Workshop on Statistical Machine Translation,
  WMT~'13, Sofia, Bulgaria, 2013, pp. 45--51.

\bibitem{machacek-bojar:2014:W14-33}
M.~Mach\'{a}\v{c}ek, O.~Bojar, Results of the {WMT14} metrics shared task, in:
  Proceedings of the Ninth Workshop on Statistical Machine Translation,
  WMT~'14, Baltimore, Maryland, USA, 2014, pp. 293--301.

\bibitem{Xavier10}
Y.~Bengio, X.~Glorot, Understanding the difficulty of training deep feedforward
  neural networks, in: Proceedings of AI \& Statistics 2010, Vol.~9, Chia
  Laguna Resort, Sardinia, Italy, 2010, pp. 249--256.

\bibitem{Doddington:2002:AEM}
G.~Doddington, Automatic evaluation of machine translation quality using n-gram
  co-occurrence statistics, in: Proceedings of the Second International
  Conference on Human Language Technology Research, HLT '02, Morgan Kaufmann
  Publishers, San Francisco, California, USA, 2002, pp. 138--145.

\bibitem{Snover06astudy}
M.~Snover, B.~Dorr, R.~Schwartz, L.~Micciulla, J.~Makhoul, A study of
  translation edit rate with targeted human annotation, in: Proceedings of the
  7th Biennial Conference of the Association for Machine Translation in the
  Americas, AMTA '06, Cambridge, Massachusetts, USA, 2006.

\bibitem{Denkowski2011}
M.~Denkowski, A.~Lavie, Meteor 1.3: Automatic metric for reliable optimization
  and evaluation of machine translation systems, in: Proceedings of the Sixth
  Workshop on Statistical Machine Translation, WMT~'11, Edinburgh, Scotland,
  2011, pp. 85--91.

\bibitem{yih-EtAl:2011:CoNLL}
W.-t. Yih, K.~Toutanova, J.~C. Platt, C.~Meek, Learning discriminative
  projections for text similarity measures, in: Proceedings of the Fifteenth
  Conference on Computational Natural Language Learning, CoNLL~'11, Portland,
  Oregon, USA, 2011, pp. 247--256.

\bibitem{mikolov2013efficient}
T.~Mikolov, K.~Chen, G.~Corrado, J.~Dean, Efficient estimation of word
  representations in vector space, arXiv preprint arXiv:1301.3781.

\bibitem{kim:2014:EMNLP2014}
Y.~Kim, Convolutional neural networks for sentence classification, in:
  Proceedings of the 2014 Conference on Empirical Methods in Natural Language
  Processing (EMNLP), Association for Computational Linguistics, Doha, Qatar,
  2014, pp. 1746--1751.

\bibitem{hochreiter1997long}
S.~Hochreiter, J.~Schmidhuber, Long short-term memory, Neural Computation 9~(8)
  (1997) 1735--1780.

\bibitem{Kalchbrenner14}
N.~Kalchbrenner, E.~Grefenstette, P.~Blunsom, A convolutional neural network
  for modelling sentences, in: Proceedings of the 52nd Annual Meeting of the
  Association for Computational Linguistics, 2014, pp. 655--665.

\bibitem{Schuster:1997}
M.~Schuster, K.~K. Paliwal, Bidirectional recurrent neural networks, IEEE
  Transactions on Signal Processing 45~(11) (1997) 2673--2681.

\bibitem{hinton:dropout}
N.~Srivastava, G.~Hinton, A.~Krizhevsky, I.~Sutskever, R.~Salakhutdinov,
  Dropout: A simple way to prevent neural networks from overfitting, Journal of
  Machine Learning Research 15 (2014) 1929--1958.

\bibitem{Hopkins2011}
M.~Hopkins, J.~May, Tuning as ranking, in: Proceedings of the 2011 Conference
  on Empirical Methods in Natural Language Processing, EMNLP~'11, Edinburgh,
  Scotland, UK., 2011, pp. 1352--1362.

\bibitem{Wang2012}
M.~Wang, C.~Manning, Spede: Probabilistic edit distance metrics for mt
  evaluation, in: Proceedings of the Seventh Workshop on Statistical Machine
  Translation, Montr{\'e}al, Canada, 2012, pp. 76--83.

\bibitem{chen-kuhn-foster:2012:WMT}
B.~Chen, R.~Kuhn, G.~Foster, Improving amber, an mt evaluation metric, in:
  Proceedings of the Seventh Workshop on Statistical Machine Translation,
  Montr{\'e}al, Canada, 2012, pp. 59--63.

\bibitem{machavcek-bojar:2011:WMT}
M.~Mach\'{a}\v{c}ek, O.~Bojar, Approximating a deep-syntactic metric for mt
  evaluation and tuning, in: Proceedings of the Sixth Workshop on Statistical
  Machine Translation, Edinburgh, Scotland, 2011, pp. 92--98.

\bibitem{Spearman}
C.~Spearman, The proof and measurement of association between two things, The
  American Journal of Psychology 15~(1) (1904) 72--101.

\bibitem{yu-etal:2015:WMT}
H.~Yu, Q.~Ma, X.~Wu, Q.~Liu, Casict-dcu participation in wmt2015 metrics task,
  in: Proceedings of the Tenth Workshop on Statistical Machine Translation,
  Association for Computational Linguistics, Lisbon, Portugal, 2015, pp.
  417--421.

\bibitem{stanojevic-simaan:2014:W14-33}
M.~Stanojevi\'{c}, K.~Sima'an, {BEER: BEtter Evaluation as Ranking}, in:
  Proceedings of the Ninth Workshop on Statistical Machine Translation,
  Baltimore, Maryland, USA, 2014, pp. 414--419.

\bibitem{wu-yu-liu:2014:W14-33}
X.~Wu, H.~Yu, Q.~Liu, {RED, The DCU-CASICT Submission of Metrics Tasks}, in:
  Proceedings of the Ninth Workshop on Statistical Machine Translation,
  Baltimore, Maryland, USA, 2014, pp. 420--425.

\bibitem{Pearson}
K.~Pearson, Notes on regression and inheritance in the case of two parents,
  Proceedings of the Royal Society of London~(58) (1895) 240--242.

\bibitem{gautam-bhattacharyya:2014:W14-33}
S.~Gautam, P.~Bhattacharyya, {LAYERED}: Metric for machine translation
  evaluation, in: Proceedings of the Ninth Workshop on Statistical Machine
  Translation, Baltimore, Maryland, USA, 2014, pp. 387--393.

\end{thebibliography}
